\documentclass[10pt,twocolumn,letterpaper]{article}

\usepackage[pagenumbers]{cvpr} 

\usepackage{amsmath,amsthm}
\usepackage{multirow}
\usepackage{wrapfig}
\usepackage{graphicx}
\newtheorem{lemma}{Lemma}
\usepackage{xcolor}
\newtheorem{theorem}{Theorem}
\definecolor{cvprblue}{rgb}{0.21,0.49,0.74}
\usepackage[pagebackref,breaklinks,colorlinks,allcolors=cvprblue]{hyperref}

\def\paperID{10316} 
\def\confName{CVPR}
\def\confYear{2025}

\title{LongDiff: Training-Free Long Video Generation in One Go}
   
\author{
Zhuoling Li$^{1}$ \quad Hossein Rahmani$^{1}$ \quad Qiuhong Ke$^{2}$ \quad Jun Liu$^{1}$\thanks{Corresponding author} \\
$^1$ Lancaster University \quad $^2$ Monash University\\
{\tt\small \{z.li81, h.rahmani\}@lancaster.ac.uk, Qiuhong.Ke@monash.edu, j.liu81@lancaster.ac.uk}
} 

\begin{document}
\maketitle
\begin{abstract}
    Video diffusion models have recently achieved remarkable results in video generation. Despite their encouraging performance, most of these models are mainly designed and trained for short video generation, leading to challenges in maintaining temporal consistency and visual details in long video generation. In this paper, 
    we propose LongDiff, a novel training-free method consisting of carefully designed components \--- Position Mapping (PM) and Informative Frame Selection (IFS) \--- to tackle two key challenges that hinder short-to-long video generation generalization: temporal position ambiguity and information dilution. Our LongDiff unlocks the potential of off-the-shelf video diffusion models to achieve high-quality long video generation in one go. 
    Extensive experiments demonstrate the efficacy of our method.
\end{abstract}

\section{Introduction}
\label{sec:intro}
With the popularity of generative AI, text-to-video generation, which aims to create video content aligned with provided textual prompts, has become a very important and hot research topic~\cite{foo2023ai}. In recent years, extensive research has been conducted in this area, where video diffusion models \cite{blattmann2023align, ho2022video, he2022latent, wang2023lavie, girdhar2023emu, khachatryan2023text2video} have made remarkable progress. Benefiting from training on a large collection of annotated video data, these models can generate high-quality videos that are even nearly indistinguishable from reality. 
However, despite the impressive generation quality, most existing approaches \cite{ho2022video, ho2022imagen, he2022latent, chen2023videocrafter1, wang2023lavie} are mainly designed and trained for short video generation, typically limited to fewer than 24 frames. Yet, in many real-world applications, like filmmaking \cite{totlani2023evolution}, game development \cite{colado2023using} and animation creation \cite{wang2024evolution}, long videos are commonly required. Thus, long video generation, with its valuable applications, has become an urgent problem to address, attracting significant attention.

To enable long video generation, a straightforward approach is to re-train video models on long-video datasets.
For example,  the works of \cite{bao2024vidu, skorokhodov2022stylegan, tian2024videotetris} directly train models to generate entire long videos by increasing computational resources and extending model capacities. Additionally, some methods adopt auto-regressive \cite{videoworldsimulators2024, henschel2024streamingt2v} or hierarchical paradigms \cite{yin2023nuwa, ge2022long}, which break down the complex task of long video generation into manageable processes by focusing on generating individual frames or short clips that are logically assembled to form long videos \cite{li2024survey}. However, these training-based methods tend to be resource and time intensive due to the complexity of their models. Moreover, the scarcity of long video datasets makes it hard to meet training needs, thus making it challenging for researchers to obtain optimal parameters for long video generation \cite{li2024survey}.

Considering the amazing performance of the off-the-shelf short video generation models, another approach is to directly adapt these models for long video generation without any parameter updates. Such a training-free approach requires neither annotated long video data nor computational resources for re-training, and thus has gained much attention, with several attempts made \cite{wang2023gen, qiu2023freenoise, lu2024freelong}. To extend videos smoothly, some of them \cite{wang2023gen, qiu2023freenoise} split the entire long video to be generated, into several overlapping short clips via a sliding window, and then process the clips asynchronously. While these methods can maintain consistency within individual video clips, the sliding window scheme limits long-range interactions between longer-distance frames, potentially leading to a lack of global temporal consistency. Another method \cite{lu2024freelong} generates the long video in one go (i.e., without using sliding-windows), and improves video quality by mixing spatial and temporal information in the frequency domain. These training-free methods that mainly rely on empirical designs, have made some progress, yet the improvements achieved still remain suboptimal. In this work, we also seek to unlock the inherent potential of the off-the-shelf short video models, to generate high-quality long videos in one go in a training-free manner. 

\begin{figure}[t!]
    \centering
     \includegraphics[width=1\linewidth]{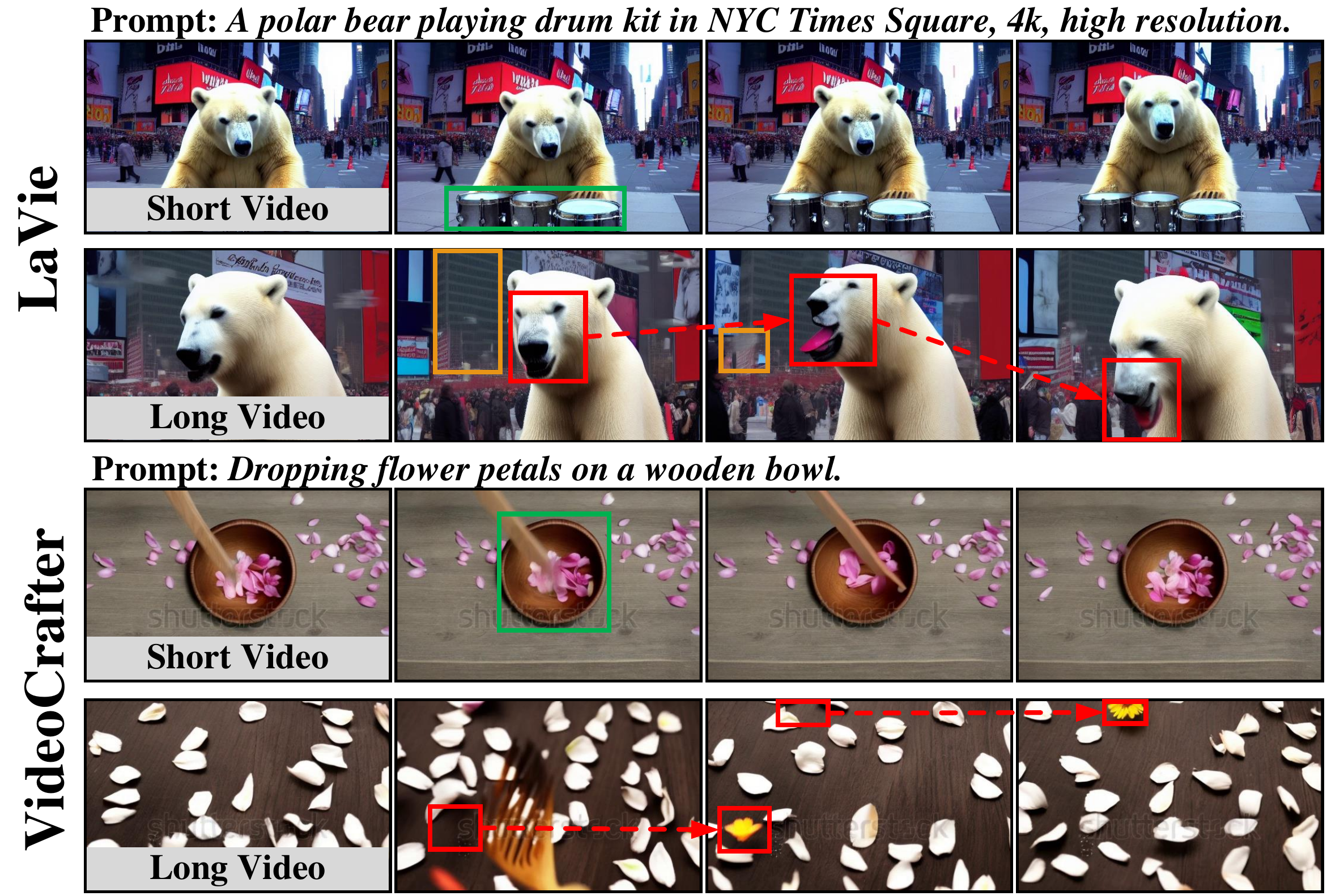}
     \vspace{-0.6cm} 
     \caption{Results of directly applying short-video diffusion models (LaVie \cite{wang2023lavie} and VideoCrafter \cite{chen2023videocrafter1} ) for long video generation. Since the spatial transformer layers in these short video models operate independently of video lengths, and the temporal transformer layers can process input sequences of various length, we only need to extend the length of the noise sequences used as the starting point for denoising to achieve long video generation. It can be observed that though short videos have good quality, long videos exhibit inferior temporal consistency (marked by the \textcolor{red}{red  boxes}), such as abrupt transitions of the polar bear's expression and the sudden appearance and disappearance of the yellow flower. 
     Additionally, the long videos lack some key visual details (marked by the \textcolor{orange}{orange boxes}), such as the missing ``\textit{drum}" and ``\textit{wooden bowl}", and blurred ``\textit{NYC Times Square}".}
     \label{fig:intro}
    \vspace{-0.3cm}
  \end{figure}

To generate videos with temporal coherence, it is crucial to capture the temporal correlations over video frames. To this end, temporal convolution and temporal transformer are incorporated in previous works \cite{wang2023modelscope, wang2023lavie, chen2023videocrafter1} to capture positional relationships over frames. 
Among them, temporal transformer with relative positional encoding techniques \cite{su2024roformer, shaw2018self} becomes more and more popular, and has been widely used in recent video diffusion models \cite{wang2023lavie, chen2023videocrafter1, yang2024cogvideox, ho2022video, he2022latent, ho2022imagen}. In this paper, we mainly focus on this category of video diffusion models. However, despite the strong sequence modeling capability of temporal transformer, as shown in \cref{fig:intro}, directly using short video models for long video generation yields low-quality results with inferior temporal consistency (e.g., abrupt transitions between frames) and a lack of visual details (e.g., blurred textures and missing critical details).

Drawing on ~\cite{han2024lm} and leveraging theoretical insights from pseudo-dimensions and information entropy, these limitations in long video generation can be further attributed to two fundamental challenges:
\textbf{(1) Temporal Position Ambiguity}.   The pseudo-dimensions-based analysis  reveals that as the length of the generated video increases, video models struggle to accurately differentiate between the relative positions of frames. This positional ambiguity disrupts the model's ability to maintain frame order, leading to compromised temporal consistency (e.g., abrupt transitions of the polar bear’s expression as illustrated in \cref{fig:intro}).
\textbf{(2) Information Dilution}. 
The information entropy-based analysis shows that 
as the length of the video sequence increases, the temporal correlation entropy between frames shows an upward trend. This indicates a reduction in per-frame information content during generation, resulting in missing visual details and reduced visual quality in long video outputs (e.g., the missing ``\textit{drum}" and ``\textit{wooden bowl}", and blurred ``\textit{NYC Times Square}" as illustrated in \cref{fig:intro}).

Based on the above analysis, these two challenges can be effectively mitigated through targeted, minor modifications to the temporal transformer layers in video diffusion models. We propose \textbf{LongDiff}, a simple yet effective method that unlocks the inherent long-video generation capability of pretrained short-video diffusion models in a training-free manner, enabling the generation of high-quality long videos with global temporal consistency and visual details. 
Our LongDiff consists of two key components: \textbf{P}osition \textbf{M}apping (PM) and \textbf{I}nformative \textbf{F}rame \textbf{S}election (IFS). PM is designed to ensure that frame positions are accurately differentiated by video models. It maps 
large numbers of distinct relative frame positions to a manageable range, preserving their distinctiveness through simple \texttt{GROUP} and \texttt{SHIFT} operations. This mapping addresses the issue of temporal position ambiguity and ensures that the video model maintains the correct frame order of the generated sequence, thereby enhancing temporal consistency. To address the issue of information dilution, the IFS strategy limits temporal correlation of each frame to its neighbor frames and a set of selected key frames. This reduces the risk of excessive entropy of temporal correlation during generation, which can impair the capture of fine details. Both PM and IFS are elegant, with minor modifications to the original temporal transformer layers, which facilitate long video generation in one go.

Our contributions are as follows. 
We propose a novel training-free method, LongDiff, consisting of two carefully designed components: position mapping and informative frame selection, to mitigate two key challenges that hinder high-quality long video generation. Our LongDiff enables pretrained short video models to generate high-quality long videos in one go and achieves state-of-the-art performance on the evaluated benchmarks.

\section{Related Work}

\noindent \textbf{Text-to-Video Diffusion Models}
are mainly constructed based on text-to-image diffusion models \cite{ramesh2021zero, saharia2022photorealistic}, incorporating temporal modules to establish correlations over video frames. Since frame order (i.e., positional relationships over video frames) is essential for maintaining temporal coherence in generated videos, some methods \cite{wang2023modelscope, chen2024videocrafter2} introduce temporal convolution to capture positional relationships of frames within a local range. However, temporal convolution may struggle with building long-range temporal correlations over frames \cite{wen2022transformers}.
To this end, many recent methods \cite{blattmann2023align, chen2023videocrafter1, wang2023lavie, girdhar2023emu, khachatryan2023text2video} also incorporate temporal transformers equipped with positional encoding techniques \cite{su2024roformer, shaw2018self,vaswani2017attention} to capture precise positional relationships over frames, facilitating temporally coherent generation. Among these, relative positional encodings \cite{su2024roformer, shaw2018self}, which can effectively capture inter-token relationships in LLMs, have been widely adopted in recent video diffusion models \cite{chen2023videocrafter1, wang2023lavie, girdhar2023emu, ho2022video}, achieving impressive video generation results.
In this paper, we mainly focus on video models that incorporate temporal transformers with relative positional encodings. 
Specifically, we aim to tackle two key challenges in long video generation and propose LongDiff, a training-free method to adapt these short video models to generate high-quality long videos.

\noindent \textbf{Long Sequence Generalization} 
is a common problem arising in various AI tasks, such as automatic speech recognition \cite{zhang2023google, moritz2019triggered}, long text comprehension \cite{chen2023extending, xiao2023efficient, han2024lm, jin2024llm}, human activity analysis \cite{zhang2020few, ahn2023star, Qu_2024_CVPR, foo2024action, Foo_2023_CVPR}, and long video generation \cite{harvey2022flexible, wang2023gen, lu2024freelong, qiu2023freenoise}.
Among these tasks, long video generation is particularly challenging due to the need to maintain both temporal content consistency and visual details throughout the sequence.
To achieve high-quality long video generation, many recent advancements focus on improving visual quality using diffusion-based techniques \cite{harvey2022flexible, voleti2022mcvd, he2022latent, yin2023nuwa}. Nuwa-XL \cite{yin2023nuwa} employs a parallel diffusion process, while StreamingT2V \cite{henschel2024streamingt2v} uses an auto-regressive approach with a short-long memory block to enhance the consistency of long video sequences. Despite their effectiveness, these methods require substantial computational resources and large-scale datasets for training.

Recently, training-free methods \cite{wang2023gen, qiu2023freenoise, lu2024freelong} that adapt off-the-shelf short video models to generate long videos without parameter updates have gained attention.
Gen-L-Video \cite{wang2023gen} extends videos by merging overlapping sub-segments using a sliding-window method during denoising. FreeNoise \cite{qiu2023freenoise} uses rescheduled noise sequences and window-based temporal attention to improve video continuity. Freelong \cite{lu2024freelong} explores long video generation in the frequency domain, improving generation quality by mixing features of different frequencies via Fourier transform.
Differently, in this work, 
we focus on two challenges, revealed by theoretical analysis, that hinder short-to-long generalization of short video models. Motivated by the analysis, we find that simple modifications to temporal transformer can enable high-quality long video generation in one go.

\section{Preliminaries}
\label{preliminary}
\noindent \textbf{Temporal Transformer in Short Video Diffusion Models.}
Most of existing short video diffusion models \cite{blattmann2023align, ho2022video, ho2022imagen, he2022latent, chen2023videocrafter1, wang2023lavie} are built on the 3D U-Net architecture, where both spatial and temporal transformer layers play important roles in video generation. 
In the spatial transformer layer, video features (i.e., hidden states of the sampled video) are processed independently over frames, 
allowing these layers to remain unaffected by the length of the video.
Given this property, we mainly focus on the temporal transformer layers,  as they play a significant role in handling the challenges of long video generation. The temporal transformer, responsible for establishing correlations over frames, is computed as follows:
\begin{equation}
\setlength{\abovedisplayskip}{3pt}
\setlength{\belowdisplayskip}{3pt}
    \label{eq:attention}
    \begin{split}
    w_{i,j} &= \frac{\exp (a_{i,j})}{\sum_{k=1}^{N} \exp (a_{i,k})}, ~~a_{i,j} = f(\mathbf{q}_i, \mathbf{k}_j, i-j)  \\
    \mathbf{o}_i &= \sum\nolimits_{j=1}^{N} w_{i,j} \mathbf{v}_j
    \end{split}
\end{equation}
where $a_{i,j}$ and $w_{i,j}$ are the attention logit and attention weight between the $i$-th query frame and the $j$-th key frame, respectively. $\mathbf{q}_i$, $\mathbf{k}_j$, and $\mathbf{v}_j$ are the query, key, and value vectors, respectively. $f(\mathbf{q}_i,\mathbf{k}_j,i-j)$ is a function that computes the attention logit by taking into account the query frame $i$ and key frame $j$, together with their relative position ($i - j$). $N$ is the number of frames in the video, and the output $\mathbf{o}_i$ is the aggregated information for the $i$-th frame. The form of $f$ varies for different relative encoding mechanisms. For simplicity, we represent \(f(\mathbf{q}_i, \mathbf{k}_j, i-j)\) as \(f(\mathbf{q}, \mathbf{k}, p)\), where the relative position of the query and key is given by $p=i-j$. Notably, when generating a video with $N$ frames, the positions of query and key frames are typically within the range $[0, N-1]$. Hence, the range of relative positions $p$ between query and key frames is $[-(N-1), N-1]$.

\section{Proposed Method}
\begin{figure}[t]
    \centering
     \includegraphics[width=1\linewidth]{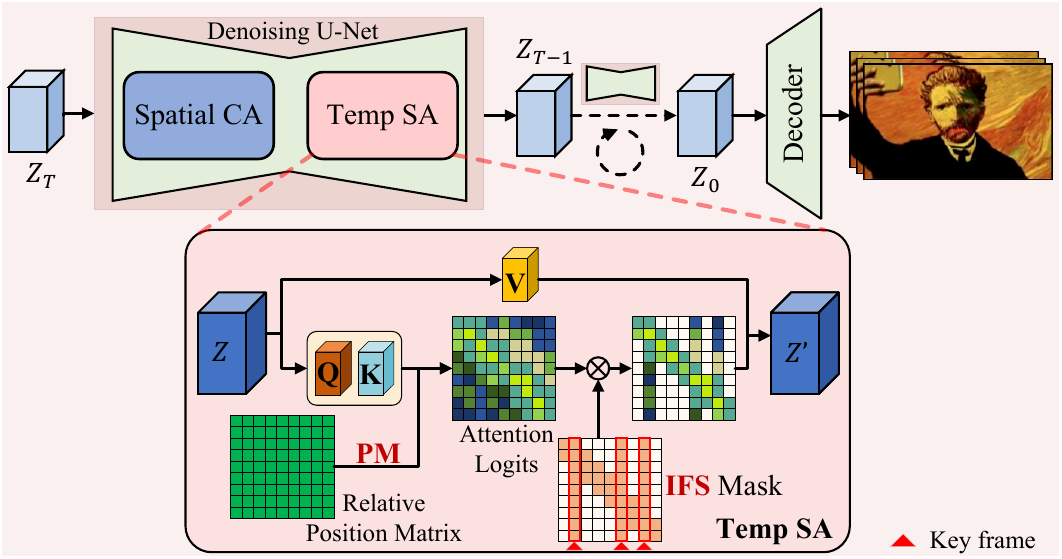}
     \vspace{-0.6cm}
     \caption{
     Overview of our proposed method. CA and SA denote cross-attention and self-attention, respectively. 
     During the denoising steps, the video hidden states \( Z \) iteratively pass through temporal transformer layers where our LongDiff mechanism is applied. LongDiff comprises two key components: Position Mapping (PM) and Informative Frame Selection (IFS), corresponding to two modifications to temporal self-attention. First, we transform the original relative position matrix (the green matrix)  via PM to alleviate the temporal position ambiguity issue. Additionally, a specially designed IFS mask restricts the temporal correlations of each query frame to both its neighbor frames and a set of detected key frames, to avoid the problem of information dilution.}
     \vspace{-0.3cm}
     \label{fig:longdiff}
  \end{figure}

We aim to adapt off-the-shelf short video models to generate high-quality long videos in one go.
To this end, 
we propose our training-free solution, LongDiff.
As shown in \cref{fig:longdiff}, LongDiff involves subtle modifications \--- Position Mapping (PM) and Informative Frame Selection (IFS) \--- to temporal attention in short video models, which are carefully designed to tackle two challenges in long video generation, namely \emph{temporal position ambiguity} and \emph{information dilution}.
Below, we elaborate on how these two challenges are revealed, and introduce how the proposed solutions address these challenges.

\subsection{Maintaining Temporal Consistency}
\label{sec:PED}

\noindent \textbf{Challenge: Temporal Position Ambiguity.}
Ensuring temporal consistency in video generation is crucial for maintaining realistic and smooth transitions between frames. This is linked to relative positional encoding (RPE), a technique used in various video generation models to encode relative positions of frames. By capturing positional differences, RPE helps models to maintain temporal coherence, guiding them to generate frames in the correct order. It is reasonable to conjecture that one reason of the loss of temporal consistency in long video generation can be  that \textit{RPEs in temporal transformer layers fail to function as intended when generating longer videos.}
This is evident in some video diffusion models \cite{chen2023videocrafter1, ho2022video}, which employ the RPE mechanism \cite{shaw2018self} to encode relative positions only within a fixed range. When generating long videos, these models assign (clip) all relative positions exceeding the maximum encoding range to the same boundary position value, which impedes the model’s ability to recognize frame order. However, even with recent RPE techniques, like RoPE \cite{su2024roformer}, which theoretically can encode relative positions in sequences significantly longer than that required for long video generation, video diffusion models \cite{wang2023lavie,yang2024cogvideox} still struggle to maintain temporal consistency in long video generation.
Inspired by \cite{han2024lm}, leveraging the pseudo-dimension technique \cite{pollard1990empirical, fioravanti2023pac, liu2024representation}, which is a metric used to assess the expressive capacity of nonlinear functions, this problem can be further investigated through the following theoretical explanation.
\begin{theorem}
    \label{the1}
Define the attention logit function in temporal attention as \(f(\mathbf{q}, \mathbf{k}, p)\),
which maps the query frame \(\mathbf{q}\), key frame \(\mathbf{k}\), and their relative position $p$ to a scalar value. 
Consider a video generation task with $N$ frames, where the model categorizes the $2N$$-$$1$ relative positions\footnote{There are $2N$$-$$1$ relative positions, because the indices of relative positions range from $-(N-1)$ to $N-1$, as discussed in Sec. \ref{preliminary}.} into \(g(N)\) groups.
Here, \(g(N) \in \mathbb{N}\) is a non-decreasing and unbounded function representing the model's capability to differentiate relative positions. Additionally, assume that any two relative positions \(p\) and \(p'\) within the same group satisfy \(d_f(p, p') \leq \epsilon\), where \(d_f\) is the distance function associated with the attention logit function \(f\). Then the following  holds:
\begin{equation}
\setlength{\abovedisplayskip}{3pt}
\setlength{\belowdisplayskip}{3pt}
\label{sup}
\sup_{-(N-1)\leq p \leq N-1} |f(\mathbf{q}, \mathbf{k}, p)| \geq \left(\frac{g(N)}{2}\right)^{\frac{1}{2r}}\frac{ \epsilon}{4e}
\end{equation}
where $r$ is the pseudo-dimension of the function class $\mathcal{H} = \{f(\cdot, \cdot, p) \mid p \in \mathbb{Z}\}$, and $e$ is the Euler's number. 
\end{theorem}

For a detailed proof of this theoretical explanation, please refer to Supplementary. 
\cref{the1} illustrates that video models' ability to distinguish between different relative positions (i.e., $g(N)$) is limited by the supremum of the temporal attention logits.
Notably, to ensure that the video is generated in the correct frame order, the ideal video model should satisfy \( g(N) = 2N - 1 \). Otherwise, at least one pair of distinct relative positions will be indistinguishable by the model. In other words, there will exist \((p, p^\prime)\) (where \( p \neq p^\prime \) and \( -(N-1) \leq p, p^\prime \leq N-1 \)) such that \( d_f(p, p^\prime) \leq \epsilon \), i.e., there will be temporal position ambiguity. Therefore, as the video sequence length \( N \) increases, we expect \( g(N) \) to also  increase accordingly in order to ensure position distinguishability. This necessitates larger supremum of the attention logits to satisfy the inequality in \cref{sup}. However, through experiments (with details in Supplementary), we observe in long video generation, e.g., videos with 128 frames, less than 40\% (the percentage is even smaller for longer videos)  of the query and key features satisfy Eq. \eqref{sup} when substituting $g(N)$ with $2N-1$, which is the requirement for the model to correctly identify frame order. This indicates the limited ability of using off-the-shelf short video models when distinguishing large numbers of distinct relative positions in long sequences, inevitably leading to degenerated temporal consistency for long video generation.

\noindent \textbf{Solution: Position Mapping (PM).}
According to \cref{the1}, an effective way to alleviate temporal position ambiguity is by limiting the number of distinct relative positions processed by the model. This can be achieved through a simple clipping operation~\cite{han2024lm}, which sets a position threshold and maps all relative positions exceeding it to the threshold value. However, this method prevents the model from distinguishing between different relative positions larger than this threshold, limiting its ability to establish correlations over distant frames. Moreover, \cref{the1} also implies that position interpolation \cite{chen2023extending}, which down-scales input relative position indices to fit within the pretraining position range, remains suboptimal, because it essentially introduces more relative positions within a short range. This requires the model to enhance its ability to distinguish positions, which is challenging in a training-free setting.

To tackle the temporal position ambiguity issue, here we propose a method that divide the relative positions in long video sequences into several position groups. Instead of using the original positions, we use the group indices to reference the corresponding positional encodings for temporal attention computation. 
This \texttt{GROUP} operation reduces the need to manage an excessive number of indistinguishable relative positions by mapping them to a smaller set of indices, while still approximately preserving the overall positional relationships over the video sequence.
However, the model still cannot distinguish relative positions within each group. 
Hence, we further design a \texttt{SHIFT} operation to recover fine-grained position distinguishability. Notably, our solution is compatible with video diffusion models that unitize different RPE techniques.
\begin{figure}[t]
    \centering
     \includegraphics[width=1\linewidth]{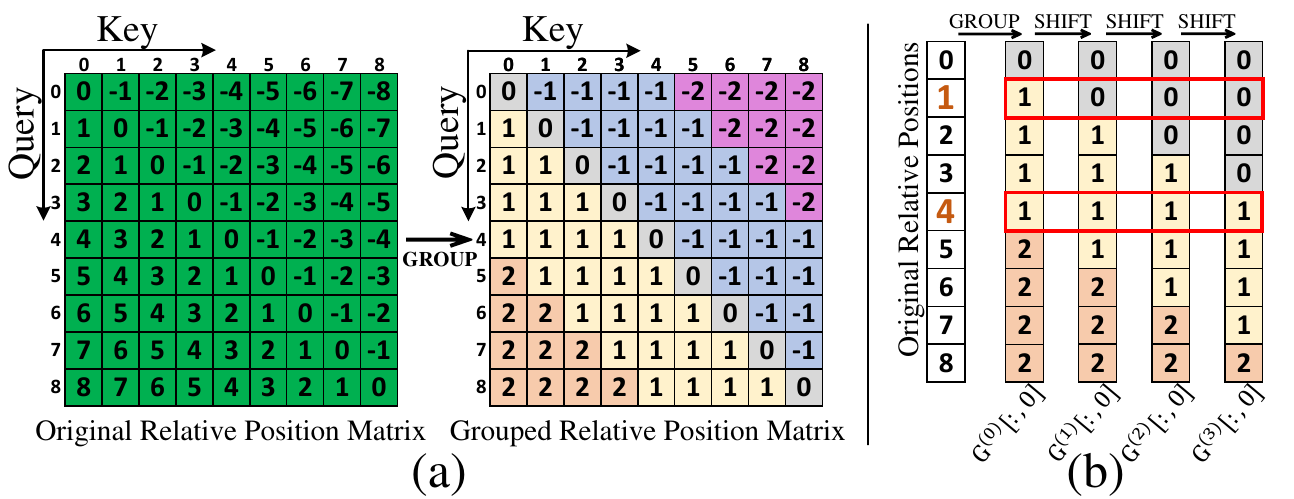}
     \vspace{-0.7cm}
     \caption{Figure (a) shows the \texttt{GROUP} operation where $N=9$ and $G=3$. The query-axis and key-axis of the matrices represent positions of the query frames and key frames, respectively. 
     Each matrix entry represents the relative position between the query and key frames. In the grouped relative position matrix, the 17 original relative positions (ranging from $-8$ to $8$) are grouped into 5 groups (from $-2$ to $2$). Figure (b) shows a simple case of the \texttt{SHIFT} operation on the first column of the shifted relative position matrix \( \mathbf{G}^{(m)} \). The red box represents the ``assignment record''.}
     \label{fig:group_shift}
     \vspace{-0.3cm}
  \end{figure}

\noindent \textbf{(1) Position Grouping.}
For a long video with $N$ frames, we map the $2N - 1$ relative positions (from $-(N - 1)$ to $N - 1$) into $2G-1$ groups, with group indices spanning from $-(G - 1)$ to $G - 1$, where $G$ is a hyperparameter satisfying $G<=N$. 
The \texttt{GROUP} operation is formulated as:
\begin{equation}
\setlength{\abovedisplayskip}{3pt}
\setlength{\belowdisplayskip}{3pt}
\label{eq:group}
    p_{g}  =
\begin{cases} 
\lceil \frac{p}{\lceil\frac{N-1}{G-1}\rceil} \rceil, & \text{if } p \geq 0, \\
\lfloor \frac{p}{\lceil\frac{N-1}{G-1}\rceil} \rfloor, & \text{if } p < 0.
\end{cases}
\end{equation}
where $p_{g}$ denotes the group index, and $p$ is the original relative position between the query and key frames. $\lceil \cdot \rceil$ and $\lfloor \cdot \rfloor$ denote the ceil and floor operations, respectively.
Notably, except for the 0-\textit{th} and the last groups, each group encapsulates $S=\lceil\frac{N-1}{G-1}\rceil$ original relative positions. The group indices are then used as the positional encoding indices in the temporal attention computation. 
A visual example of the \texttt{GROUP} operation is provided in \cref{fig:group_shift}(a), where $N = 9$ and $G = 3$.

\noindent
\textbf{(2) Position Shifting.}
While the \texttt{GROUP}  enables the model to manage a reduced set of relative positions, it introduces ambiguity in frame order, as the model cannot discern the relative positions within the same position group. For example, as shown in \cref{fig:group_shift}(a), original relative positions $[1,2,3,4]$ are now all mapped into the 1-\textit{th} group, and $[1,1,1,1]$ are then used as their actual relative positions. To tackle this problem, we recover position distinguishability within each group through a  tailored \texttt{SHIFT} operation. This \texttt{SHIFT} operation is applied to the grouped position matrix (denoted as $\mathbf{G}^{(0)} \in \mathbb{R}^{N \times N}$) obtained through the \texttt{GROUP} operation.  
We denote the position matrix after the $m$-\textit{th} shift as $\mathbf{G}^{(m)}$. To simplify the explanation of the \texttt{SHIFT} operation, we take the entries in the first column of $\mathbf{G}^{(0)}$ as an example. As shown in \cref{fig:group_shift}(b), given $\mathbf{G}^{(0)}[:,0]$, which contains 3 groups  (namely, $0$, $1$ and $2$) mapped from the original relative positions, we apply a downward shift (padding zeros on the top) to the entries in $\mathbf{G}^{(0)}[:,0]$. After each \texttt{SHIFT} operation, each entry is assigned a relative position, which may differ from its initial one. After three \texttt{SHIFT} operations, each entry accumulates four position assignments (including the initial grouped position). For example, the second entry (with original position value of 1) has an assignment record of $[1,0,0,0]$, while the fifth entry (with original position value of 4) has $[1,1,1,1]$. In this case, each entry thereby receives a unique ``assignment record'', where the sum of the record values for each entry exactly corresponds to its original position, as shown in the red boxes in \cref{fig:group_shift}(b). This assignment record allows us to distinguish entries even within the same group. As we aim for a training-free approach, we leverage the distinctness of the assignment record to distinguish each entry by separately computing temporal attention using its assigned positions and then simply average the resulting softmax attentions. Below we introduce the details of how the \texttt{SHIFT} operation is applied to the entire position matrix $\mathbf{G}^{(m)}$. 
\begin{figure}[t] 
    \centering
     \includegraphics[width=1\linewidth]{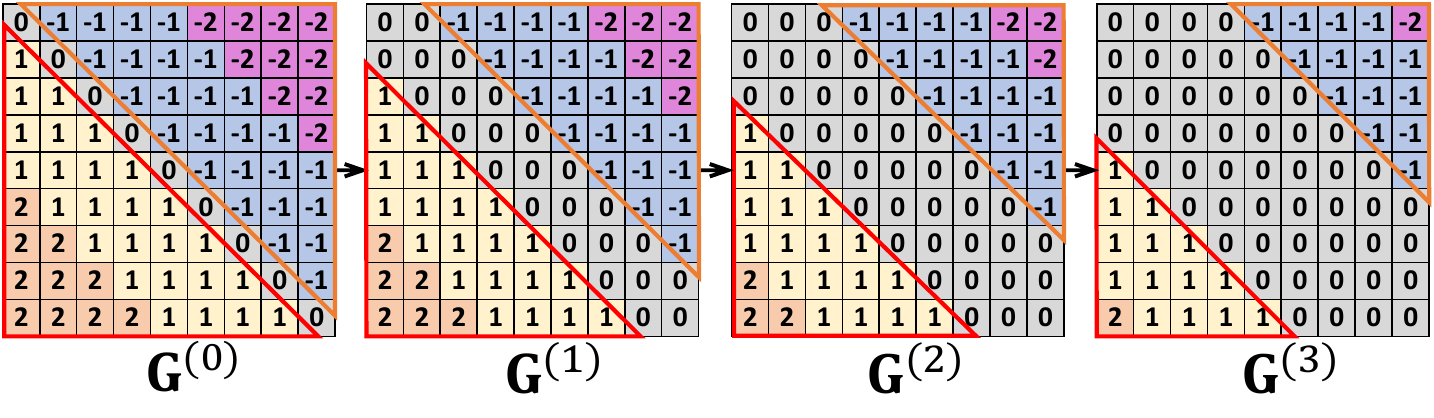}
     \vspace{-0.75cm}
     \caption{Illustration of the \texttt{SHIFT} operation. In each \texttt{SHIFT} operation, each entry in the upper triangle is shifted to the right by one position, with zeros added at the left. Meanwhile, each entry in the lower triangle is shifted downward by one position, with zeros added at the top.}
     \label{fig:shift}
     \vspace{-0.4cm}
  \end{figure}

As shown in \cref{fig:group_shift}(a), relative position matrices are always anti-symmetric, which exhibit beneficial properties, namely, the relative position of a frame with itself is always 0, and for any two frames, swapping their positions results in the relative position value being negated. Hence, when shifting a certain entry in such an anti-symmetric position matrix, the entry at its symmetric position should also change accordingly. With this insight, as shown in \cref{fig:shift}, in each \texttt{SHIFT} operation, we keep the diagonal entries as 0 and shift each entry in the lower triangle downward by one position. Additionally, we simultaneously update the corresponding entries in the upper triangle to ensure $\mathbf{G}^{(m)}_{i,j}= -\mathbf{G}^{(m)}_{j,i}$ due to the anti-symmetric property, which can be achieved through shifting entries row-wise in the upper triangle. Our \texttt{SHIFT} operation is formulated as:
\begin{equation}
\setlength{\abovedisplayskip}{3pt}
\setlength{\belowdisplayskip}{3pt}
    \mathbf{G}_{i,j}^{(m+1)} = 
    \begin{cases} 
        \mathbf{G}_{i,j-1}^{(m)}, & \text{if } i<j, \\
        \mathbf{G}_{i,j}^{(m)}, & \text{if } i=j, \\
        \mathbf{G}_{i-1,j}^{(m)}, & \text{if } i>j. \\
    \end{cases}
\end{equation}
Notably, the number of required shifts ($M$) is determined by the size of each position group ($S$), and is given by $M=S-1$.
Then, we use the \( M+1 \) relative position matrices \( \{\mathbf{G}^{(m)}\}_{m=0}^M \) to compute the temporal attention, respectively. We average the resulting softmax-attentions as the final output of the temporal attention:
\begin{equation}
\setlength{\abovedisplayskip}{3pt}
\setlength{\belowdisplayskip}{3pt}
    \label{eq:ave}
    \resizebox{0.90\hsize}{!}{$
    \mathbf{A}_{i,j}\! =\! \frac{1}{M\!+1}\sum_{m=0}^{M} \mathbf{A}_{i,j}^{(m)},~
    \mathbf{A}^{(m)}_{i,j} = \frac{\exp({f(\mathbf{q}_i, \mathbf{k}_j, \mathbf{G}_{i,j}^{(m)})})}{\sum_{k} \exp{(f(\mathbf{q}_i, \mathbf{k}_{k}, \mathbf{G}_{i,k}^{(m)}))}}$}
\end{equation}
where \( \mathbf{A} \in \mathbb{R}^{N \times N} \) is the final temporal attention output, and \( f \) is the function described in \cref{eq:attention} that computes the attention logit, taking the relative position into account. 

Through the aforementioned \texttt{GROUP} and \texttt{SHIFT} operations, PM maps large numbers of distinct relative positions into several position groups while maintaining their distinctness. 
It is worth noting that the temporal attention computation is only a part of the entire denoising process, and the $M$ \texttt{SHIFT} operations can be performed in parallel. Thus, our PM results in only a minor decrease in inference speed.

\subsection{Maintaining Visual Details}
\noindent  \textbf{Challenge: Information Dilution.}
\label{sec:dilution}
As shown in \cref{fig:intro}, aside from inferior temporal consistency, 
directly generating long videos using short video models results in a lack of visual details, including blurred textures and missing critical details. 
This problem can be further investigated with the following theoretical explanation, which is obtained by interpreting the information passing mechanism of the temporal transformer from the perspective of information entropy~\cite{han2024lm}:
\begin{theorem}
\label{t2}
    When generating a video with $N$ frames, the information entropy $H$ of temporal correlations over frames of the video sequence, is lower bounded by:
\begin{equation}
\setlength{\abovedisplayskip}{3pt}
\setlength{\belowdisplayskip}{3pt}
    \label{eq:entropy}
    H\left(\frac{e^{a_i}}{\Sigma^{N}_{j=1}e^{a_j}}|1\leq i\leq N\right)  \geq \ln N -2B,
\end{equation}
where $\{a_i\}_{i=1}^N$ are the attention logits with boundary $[-B, B]$.
\end{theorem}

For a detailed proof of \cref{t2}, please refer to  Supplementary. 
\cref{eq:entropy} implies that as the video length $N$ increases, the information entropy $H$ of the temporal correlations over frames exhibits an increasing trend. This thus indicates a decrease in the effective information carried by each frame during video generation, which may cause the model missing detailed information in some key frames and finally resulting a lack of visual details in long videos.

\noindent \textbf{Solution: Informative Frame Selection (IFS).}
Based on \cref{t2}, the information dilution issue in long video generation is related to the number of frames involved in information passing during attention. Hence, it seemingly can be alleviated by using a fixed-length attention window to restrict information passing to neighbor frames \cite{qiu2023freenoise, han2024lm}.
This approach indeed maintains visual details, which however hinders long-range interactions between remoter frames, leading to a lack of global temporal consistency in the generated videos. Different from this approach \cite{qiu2023freenoise, han2024lm}, we propose to limit the temporal correlations for each frame to both its neighbor frames and a set of selected key frames that serve as a summary of the entire video. This design only uses informative frames for information exchange, reducing the risk of excessive information entropy of temporal correlations during generation, while maintaining global consistency by capturing the entire video’s information through selected key frames.

Specifically, to select frames containing important information for video generation, we exploit a simple yet effective key-frame detection mechanism \cite{dirfaux2000key} to identify frames that may reflect scene changes and significant events in the video.
Additionally, because no real video data exists until the generation process is complete, we convert the input sequential features of each temporal transformer layer into a pseudo-video to ensure compatibility with our key-frame detection pipeline.

\noindent \textbf{(1) Pseudo-Video Construction.}
The input video feature \( F \) (i.e., hidden states of the sampled video) of each temporal transformer layer has dimensions \( \mathbb{R}^{N \times C \times hw} \), where \( N \) is the number of video frames, \( h \) and \( w \) are the sizes of each frame’s feature map, and \( C \) is the number of channels.
To make \( F \) suitable for the existing key-frame detection pipeline, we perform max-pooling, average-pooling, and min-pooling operations along the channel dimension, and then stack these 3 pooled features along the channel dimension. This yields a down-sampled video feature  \( F^\prime \in \mathbb{R}^{N \times 3 \times hw } \), reserving essential semantic information from the original video feature while matching the input shape required by the key-frame detection pipeline.
We further normalize and map values of $F^\prime$ to integer values within the range \([0, 255]\), simulating real video  data. We denote the resulted pseudo-video as \( V \):
\begin{equation}
\setlength{\abovedisplayskip}{3pt}
\setlength{\belowdisplayskip}{3pt}
    V = \text{round}\left(\frac{F^\prime - \min(F^\prime)}{\max(F^\prime) - \min(F^\prime)} \times 255\right)
\end{equation}

\noindent \textbf{(2) Key-Frame Detection.}
After obtaining the pseudo-video data $V$, we employ a simple and efficient key-frame detection mechanism \cite{dirfaux2000key}. Specifically, we first uniformly divide \( V \) into \( n \) video shots, and then select one key frame in each video shot. 
We introduce image entropy, which reflects the complexity and information content of an image, to assess the importance of each frame in the video shot. The image entropy is computed based on the distribution of pixel values, defined as follows:
\begin{equation}
\setlength{\abovedisplayskip}{3pt}
\setlength{\belowdisplayskip}{3pt}
    H(k) = -\Sigma_x {p(x,k) \log_2 (p(x,k))}
\end{equation}
where \( p(x,k) \) is the probability of the luminance value \( x \) in the appearance histogram of the $k$-\textit{th} frame in the video shot. 

In addition, we also consider frame differencing, which compares the pixel-wise differences between consecutive frames to identify video content changes. That is:
\begin{equation}
\setlength{\abovedisplayskip}{3pt}
\setlength{\belowdisplayskip}{3pt}
    SAD(k) = \sum\nolimits_{i,j} |I(i,j,k) - I(i,j,k-1)|
\end{equation}
where \( I(i,j,k) \) denotes the pixel value of the \( k \)-th frame at position \( (i,j) \).
The importance score of the frame is given by combining these two measures:
\begin{equation}
\setlength{\abovedisplayskip}{3pt}
\setlength{\belowdisplayskip}{3pt}
    \label{eq:score}
    Score(k) = \alpha H(k) + SAD(k) 
\end{equation}
where \( \alpha \) is a weighting factor. The frame with the highest score in each video shot is selected as a key frame. 

\noindent \textbf{(3) IFS Mask.}
Finally, we insert a specially designed mask into the temporal attention to restrict the temporal correlations of each frame to only its neighbor frames and the selected key frames. This mask is formulated as:
\begin{equation}
\setlength{\abovedisplayskip}{3pt}
\setlength{\belowdisplayskip}{3pt}
    \label{eq:mask}
    \begin{split}
        \text{Mask}_{ij} = 
        \begin{cases}
            1, & \text{if } |i-j| \leq L \text{ or } j \text{ is a key frame}, \\
            0, & \text{otherwise}.
        \end{cases}
    \end{split}
\end{equation}
where $L$ is a hyperparameter determining the range of used neighbor frames. 

\subsection{Overall Inference}

During video generation, features are iteratively passed through temporal attention layers, where LongDiff operates. Specifically, before temporal attention computation, each entry in the original relative position matrix is assigned new positions through the \texttt{GROUP} and \texttt{SHIFT} operations, resulting in new position matrices (\( \{\mathbf{G}^{(m)}\}_{m=0}^M \)). 
These position matrices are then used for temporal attention computation. In this process, the IFS mask filters out uninformative frames by making their associated softmax-attention values to 0.
The resulting softmax attention values are averaged to obtain the final temporal attention (see Eq. (\ref{eq:ave})).

\section{Experiments}

\noindent
\textbf{Implementation Details.}
To evaluate the effectiveness of our proposed LongDiff, we conduct experiments using open-source diffusion-based text-to-video generation models: LaVie \cite{wang2023lavie} and VideoCrafter-512 \cite{chen2023videocrafter1} (more models in Supplementary). 
LaVie utilizes Rotary Position Embedding (RoPE) \cite{su2024roformer} as RPE to capture relative positions across frames, while VideoCrafter-512 employs the RPE mechanism from \cite{shaw2018self}. Both models are trained to generate 16-frame short videos at a 320$\times$512 resolution. We equip them with our LongDiff to generate long videos with 128 frames. More implementation details are in Supplementary.

\noindent
\textbf{Evaluation Metrics.}
Following FreeLong \cite{lu2024freelong}, we use metrics from VBench \cite{huang2024vbench} to evaluate video quality, including Subject Consistency (SC), Background Consistency (BC), Motion Smoothness (MS), Temporal Flickering (TF), Imaging Quality (IQ), and Overall Consistency (OC). More details about these metrics are in Supplementary.
We use the same 200 test prompts from VBench \cite{huang2024vbench} as FreeLong \cite{lu2024freelong} for evaluation.

\subsection{Evaluation on Long Video Generation}

\begin{table}[t!]
    \centering
    \scalebox{0.65}{
    \begin{tabular}{lcccccc}
        \toprule
        Method & SC $\uparrow$ & BC $\uparrow$ & MS $\uparrow$  & TF $\uparrow$ & IQ $\uparrow$ &OC $\uparrow $\\
        \midrule
        VideoCrafter-512 + Direct & 88.62   & 91.86 &  85.30 & 78.53 & 65.38 & 20.71 \\
        VideoCrafter-512 + Sliding & 83.75	& 91.14	& 92.12	& 90.58	&  67.25   & 21.51      \\ 
        VideoCrafter-512 + FreeNoise\cite{qiu2023freenoise} & 91.43 &93.48 &93.33 & 91.88 &68.39 & 22.69\\
        VideoCrafter-512 + FreeLong \cite{lu2024freelong} & 90.84 & 92.37  & 89.11  &88.46& 66.62& 21.85\\
        \midrule
        VideoCrafter-512 + Our LongDiff & \textbf{93.69} & \textbf{95.59} & \textbf{94.59} &\textbf{93.35} &\textbf{70.03}&\textbf{23.17} \\
        \midrule
        LaVie + Direct &  88.95  & 93.23    & 92.77  & 91.44& 64.76 &22.34\\
        LaVie + Sliding & 85.80  & 92.83    & 95.79 & 94.00 & 66.57 &23.46\\
        LaVie + FreeNoise \cite{qiu2023freenoise} &  92.30   &95.87   &96.32    & 94.94  &67.14&24.42\\
        LaVie + FreeLong \cite{lu2024freelong} & 95.16 & 96.80   & 96.85  &96.04& 67.55&24.56 \\
        \midrule
        LaVie + Our LongDiff  & \textbf{98.10} & \textbf{98.23} & \textbf{97.46} &\textbf{96.84} &\textbf{68.83} &\textbf{25.24}\\
        \bottomrule
    \end{tabular}}
     \vspace{-0.2cm}
    \caption{Quantitative comparisons of longer video generation (128 frames) on the video models VideoCrafter-512 and LaVie. Results on more short video models are provided in Supplementary.}
    \label{tab:main}
\end{table}

\begin{figure}[t]
    \centering\textbf{}
    \includegraphics[width=0.65\linewidth]{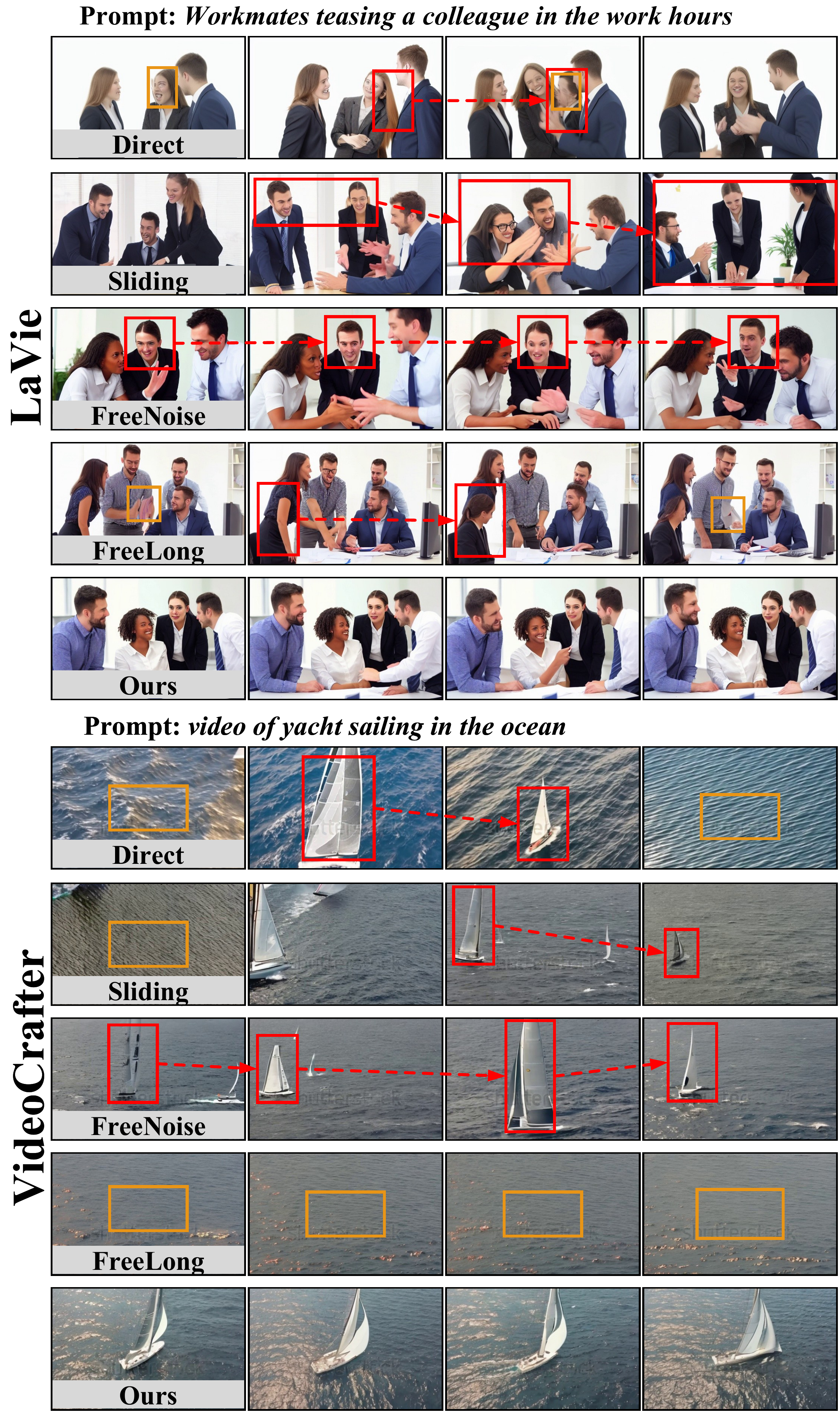}
    \vspace{-0.2cm}
     \caption{Qualitative comparisons of longer video generation. (128 frames). We illustrate inferior temporal consistency and the lack of visual details using \textcolor{red}{red} and \textcolor{orange}{orange} boxes, respectively. More examples are in Supplementary.}
\vspace{-0.2cm}
 \label{fig:longvis}
\end{figure}

We compare our LongDiff with other training-free diffusion-based long video generation methods \cite{qiu2023freenoise, lu2024freelong} and two basic methods, called \textbf{Direct} and \textbf{Sliding}. In Direct, long videos are generated directly from short video models by extending the initial noise sequence as the starting point for denoising. In Sliding, we adopt temporal sliding windows \cite{qiu2023freenoise} to process a fixed number of frames at a time. We generate videos with 128 frames for comparison. \cref{tab:main} displays the comparative results. We observe that videos generated by Direct suffer significantly in quality due to short-to-long generalization challenges. Compared to Direct, the methods of Sliding, FreeNoise, and FreeLong achieve better performance. Among them, Sliding and FreeNoise improve video quality by employing sliding-windowed attention to build temporal correlations only within neighbor frames.  
On the other hand, FreeLong improves long video generation through fusing spatial and temporal information in the frequency domain.
Compared to these methods, our method achieves the best performance across all metrics, demonstrating the efficacy of the proposed LongDiff. We show qualitative comparisons in \cref{fig:longvis}.

\subsection{Ablation Studies}
In this section, we equip the short video model LaVie \cite{wang2023lavie} with our LongDiff for analysis. By default, the short video model is trained on 16-frame videos, and we adapt it to generate 128-frame videos. 
More results are in Supplementary.
\setlength{\columnsep}{0.05in}
\begin{wraptable}[6]{r}{0.69\columnwidth}
\vspace{-0.3cm}
\centering
\resizebox{\linewidth}{!}{
\begin{tabular}{lcccccc}
    \toprule
    Method & SC $\uparrow$ & BC $\uparrow$ & MS $\uparrow$  & TF $\uparrow$ & IQ $\uparrow$&OC $\uparrow$  \\
    \midrule
    Ours (w/o PM)  &  91.85& 94.79& 95.12& 93.26& 65.73& 22.58\\
    Ours (w/o IFS) &  94.43&96.37&93.65&92.85&65.45&23.46\\
    \midrule
    Ours & 98.10 & 98.23 & 97.46& 96.84 &68.83&25.24\\
    \bottomrule
\end{tabular}}
\vspace{-0.3cm}
\caption{Ablation study for main components of LongDiff.}
\label{tab:ablation_main}
\end{wraptable}
\noindent \textbf{Impact of Main Components of LongDiff}. 
First, we verify the impact of the key components of LongDiff by comparing the following variants:
1) \textbf{Ours (w/o PM)}, where we remove the \texttt{GROUP} and \texttt{SHIFT} operations and thus use the original relative positions to compute temporal attention.
2) \textbf{Ours (w/o IFS)}, where we remove the IFS mask in temporal attention computation, requiring each query frame to correlate with all frames for information passing during video generation.
The results in \cref{tab:ablation_main} indicate that both PM and IFS significantly contribute to LongDiff's performance. 

\setlength{\columnsep}{0.05in}
\begin{wraptable}[6]{r}{0.69\columnwidth}
\vspace{-0.3cm}
\centering
\resizebox{\linewidth}{!}{
    \begin{tabular}{lcccccc}
        \toprule
        Method & SC $\uparrow$ & BC $\uparrow$ & MS $\uparrow$  & TF $\uparrow$ & IQ $\uparrow$  &OC $\uparrow $ \\
        \midrule
        Clip & 91.02	&94.18&94.77&92.87&65.12&22.43\\
        Interpolation &92.52&95.29&	95.25&	94.76&	66.43&23.01\\
        Group         & 94.49&96.63&96.74&95.79&67.45&24.62\\
        \midrule
        Our PM & 98.10 & 98.23 & 97.46& 96.84 &68.83&25.24\\
        \bottomrule
    \end{tabular}}
\vspace{-0.3cm}
\caption{Ablation study for operations in PM.}
\label{tab:ablation_PM}
\vspace{-0.4cm}
\end{wraptable}

\noindent \textbf{Impact of Operations in Position Mapping}. 
Next, we explore the impact of operations used in PM, by comparing the following variants:
1) \textbf{Clip}, where relative positions that exceed the range encountered during pretraining are set to the maximum or minimum relative position within that range.
2) \textbf{Interpolation}, where we linearly downscale input relative positions to fit within the pretraining position range. 
3) \textbf{Group}, where we remove the \texttt{SHIFT} operation, and only apply the \texttt{GROUP} operation for position mapping. As shown in \cref{tab:ablation_PM}, our  method achieves better performance than all other variants, demonstrating the efficacy of PM that allows the model to avoid large numbers of distinct relative positions while preserving position distinguishability.

\setlength{\columnsep}{0.05in}
\begin{wraptable}[8]{r}{0.69\columnwidth}
\vspace{-0.3cm}
\centering
\resizebox{\linewidth}{!}{
\begin{tabular}{lcccccc}
        \toprule
        Method & SC $\uparrow$ & BC $\uparrow$ & MS $\uparrow$  & TF $\uparrow$ & IQ $\uparrow$&OC $\uparrow$  \\
        \midrule
        Neighbor &90.12&93.54&95.31&94.27&66.94&24.15\\
        Neighbor (Plus) &90.33&93.59&94.87&94.29&67.13&24.21\\
        Key Frame &90.29&94.21&93.20&92.33&62.83&22.12\\
        Key Frame (Plus)&90.17&94.08&93.47&92.55&63.00&22.33\\
        Neighbor+Uniform &96.33&97.09&96.51&96.29&68.20&24.85
\\
        Neighbor+Random &95.77&96.64&96.23&96.14&67.77&24.69\\
        \midrule
        Our IFS & 98.10 & 98.23 & 97.46& 96.84 &68.83& 25.24\\
        \bottomrule
    \end{tabular}}
\vspace{-0.3cm}
\caption{Ablation study for different informative frame selection mechanisms.}
    \label{tab:FS}
\end{wraptable}

\noindent \textbf{Impact of Different Informative Frame Selection Mechanisms}. 
In our LongDiff, the IFS mask is used to select informative frames for information passing in temporal attention computation. Here, we further explore different frame selection mechanisms by comparing the following variants:
1) \textbf{Neighbor}, where for each query frame, we remove the detected key frames and only select its neighbor frames following \cite{qiu2023freenoise} for temporal attention computation.
2) \textbf{Neighbor (Plus)}, where for each query frame, compared to the Neighbor variant, we additionally select $n$ (the number of detected key frames) neighbor frames.
3) \textbf{Key Frame}, where for each query frame, we remove its neighbor frames and only select the detected key frames.
4) \textbf{Key Frame (Plus)}, where for each query frame, compared to the Key Frame variant, we additionally select $2L$ (the number of neighbor frames) detected key frames.
5) \textbf{Neighbor+Uniform}, where, besides neighbor frames, we uniformly select frames at equal intervals from the entire sequence and allow each frame to attend to these selected frames for attention computation.
6) \textbf{Neighbor+Random}, where, besides neighbor frames, we randomly select frames from the entire sequence.
As shown in \cref{tab:FS}, we observe that all the variants yield inferior results. 
This suggests that both neighbor and key frames carry informative content. Our IFS approach, which selects a proper combination of these frames for information passing during video generation, achieves the best performance.

\section{Conclusion}

In this paper, we propose LongDiff, a novel training-free method that involves only minor modifications, namely position mapping and informative frame selection, to temporal transformer layers. These modifications are carefully designed to tackle two key challenges in generating long videos through short video diffusion models: \emph{temporal position ambiguity} and \emph{information dilution}.
Extensive experiments demonstrate that LongDiff significantly outperforms existing training-free methods, achieving high-quality long video generation in one go.

\noindent \textbf{Acknowledgement.} 
This research was supported by the Australian Government through the Australian Research Council's DECRA funding scheme (Grant No.: DE250100030).

\setlength{\columnsep}{0.3125in}

{
    \small
    \bibliographystyle{ieeenat_fullname}
    \bibliography{main}
}
\clearpage
\setcounter{page}{1}
\maketitlesupplementary

\appendix

\section{More Implementation Details}
In our main paper, we equip  LaVie \cite{wang2023lavie} and VideoCrafter \cite{chen2023videocrafter1} with our LongDiff to generate 128-frame (i.e., $N=128$) long videos. Following \cite{qiu2023freenoise, lu2024freelong}, during sampling, we employ the noise shuffle mechanism  and perform DDIM sampling with 50 denoising steps. We set $G$ in Eq.(3) to 16. The weighting factor $\alpha$ in Eq.(10) is set to 2.  
We set the neighbor range $L$ in Eq.(11) to 8.  In addition to the neighbor frames, we also select $n=8$ key frames for temporal attention computation.
Notably, we uniformly sample 50\% of the temporal attention layers in the short video model and replace them with our LongDiff module. 
All experiments are conducted using NVIDIA 6000 Ada GPUs.

\section{More Evaluation Metrics Details}
Following FreeLong \cite{lu2024freelong}, we use metrics from VBench \cite{huang2024vbench} to evaluate video quality.
For video consistency, we report: 1) Subject Consistency (SC), measured by DINO \cite{caron2021emerging} feature similarity across frames, to check object appearance stability, and 2) Background Consistency (BC), calculated with CLIP \cite{radford2021learning} feature similarity across frames. For video fidelity, we assess 1) Motion Smoothness (MS) using AMT \cite{li2023amt} motion priors, 2) Temporal Flickering (TF) via mean absolute difference between static frames, and 3) Imaging Quality (IQ), measured by MUSIQ \cite{ke2021musiq}. For video-text consistency, we employ Overall Consistency (OC) from ViCLIP \cite{wang2023internvid} to capture both semantic and style information.

\section{Additional Ablation Studies}
We here conduct more ablation experiments about our LongDiff based on the short video model LaVie.

\subsection{Impact of the Number of Position Groups} 
In Position Mapping (PM), we map $2N-1$ (from $-(N-1)$ to $(N-1)$) 
original relative positions into $2G-1$ groups to make the model avoid handling large numbers of distinct positions.
In our main paper, we set $G=16$ for LaVie. Here we evaluate other choices of $G$, and report the results in \cref{tab:G}. 
We find that performance improves when we increase $G$, until $G$
reaches 16, where the improvement tapers off. Thus, we set $G$ = 16.

\begin{table}[h]
\centering
\resizebox{\linewidth}{!}{
\begin{tabular}{lcccccc}
    \toprule
    Method & SC $\uparrow$ & BC $\uparrow$ & MS $\uparrow$  & TF $\uparrow$ & IQ $\uparrow$&OC $\uparrow$  \\
    \midrule
    $G=8$  &  93.47& 	95.68& 	95.72& 	94.19& 	66.53& 	23.77\\
    $G=12$  & 96.19&97.17&96.74&95.74&67.88&24.42\\
    $G=16$  & 98.10 & 98.23 & 97.46& 96.84 &68.83&25.24\\
    $G=20$  &  97.25&97.76&97.14&96.45&68.41&24.98\\
    \bottomrule
\end{tabular}}
\caption{Ablation study for the number of position groups.}
\label{tab:G}
\end{table}

\subsection{Impact of the Number of Key Frames}
In our LongDiff, we establish temporal correlations between each frame and its neighboring frames, as well as $n$ key frames selected using a key-frame detection pipeline. Here, we also evaluate other choices of $n$. 
As shown in \cref{tab:IFS-n}, the model performance reaches optimal results at $n=8$. Thus, we set $n=8$ in the experiments to achieve a good result.
\begin{table}[t]
\centering
\resizebox{\linewidth}{!}{
\begin{tabular}{lcccccc}
    \toprule
    Method & SC $\uparrow$ & BC $\uparrow$ & MS $\uparrow$  & TF $\uparrow$ & IQ $\uparrow$&OC $\uparrow$  \\
    \midrule
    $n=4$  &  92.96&95.98&96.97&94.67&67.65&24.65\\
    $n=6$  &97.19&97.52&97.18&95.74&68.11&25.01\\
    $n=8$  & 98.10 & 98.23 & 97.46& 96.84 &68.83&25.24\\
    $n=10$  &97.52&97.85&97.29&96.18&68.31&25.13\\
    \bottomrule
\end{tabular}}
\caption{Ablation study for the number of key frames.}
\label{tab:IFS-n}
\end{table}

\subsection{Impact of the Number of Neighbor Frames}
Here, we also explore the impact of using different numbers $L$ of neighbor frames for temporal attention computation. As shown in \cref{tab:IFS-L}, the model's performance reaches its highest value at $L=8$. 
We thus set $L=8$ in our experiments.

\setlength{\columnsep}{0.05in}
\begin{table}[h]
\centering
\resizebox{\linewidth}{!}{
\begin{tabular}{lcccccc}
    \toprule
    Method & SC $\uparrow$ & BC $\uparrow$ & MS $\uparrow$  & TF $\uparrow$ & IQ $\uparrow$&OC $\uparrow$  \\
    \midrule
    $L=2$  &94.11&96.18&95.28&94.54&65.77&23.61\\
    $L=4$  &96.90&97.61&96.84&96.15&67.91&24.76\\
    $L=8$  & 98.10 & 98.23 & 97.46& 96.84 &68.83&25.24\\
    $L=16$  &96.43&97.37&96.55&95.87&67.58&24.59\\
    \bottomrule
\end{tabular}}
\caption{Ablation study for the number of neighbor frames.}
\label{tab:IFS-L}
\end{table}

\begin{table}[t]
\centering
\resizebox{\linewidth}{!}{
\begin{tabular}{lcccccc}
    \toprule
    Method & SC $\uparrow$ & BC $\uparrow$ & MS $\uparrow$  & TF $\uparrow$ & IQ $\uparrow$&OC $\uparrow$  \\
    \midrule
    Max  &97.28&97.70&97.02&96.58&68.37&25.06\\
    Min  &96.71&97.33&96.71&96.40&68.33&24.93\\
    Average&97.67&97.95&97.23&96.73&68.68&25.14\\
    \midrule
    Ours  & 98.10 & 98.23 & 97.46& 96.84 &68.83&25.24\\
    \bottomrule
\end{tabular}}
\caption{Ablation study for the mechanism to downsample video features.}
\label{tab:down}
\end{table}

\subsection{Impact of the Mechanism to Down-Sample Video Features}
In our LongDiff, we use a combination of max-pooling, average-pooling, and min-pooling operations to reduce the channel dimension of the video feature $F$ to three channels, aligning it with the input shape required by the key-frame detection pipeline. Here, we evaluate the efficacy of this mechanism by comparing the following variants:
1) \textbf{Max}, where only the max-pooling operation is used, and the result is replicated three times along the channel dimension.
2) \textbf{Min}, where only the min-pooling operation is used, and the result is replicated three times along the channel dimension.
3) \textbf{Average}, where only the average-pooling operation is used, and the result is replicated three times along the channel dimension.
As shown in \cref{tab:down}, we observe that using the combination of max-pooling, average-pooling, and min-pooling operations achieves the best performance. Notably,  all of these variants of our LongDiff consistently outperform the previous state-of-the-art methods\cite{lu2024freelong,qiu2023freenoise}.

\subsection{Impact of the Weighting Factor $\alpha$ }
In our LongDiff, we use two measures—image entropy and frame differencing—to select the most important frame in each shot of the pseudo-video as a key frame. Here, we evaluate the impact of $\alpha$, which weights these two measures, and report the results in \cref{tab:a}. As shown, the model achieves the best performance with $\alpha=2$. Therefore, we set $\alpha$ to 2 in our experiments to obtain optimal results.
\begin{table}[h]
\centering
\resizebox{\linewidth}{!}{
\begin{tabular}{lcccccc}
    \toprule
    Method & SC $\uparrow$ & BC $\uparrow$ & MS $\uparrow$  & TF $\uparrow$ & IQ $\uparrow$&OC $\uparrow$  \\
    \midrule
    $\alpha=0$  & 97.10&97.58&96.92&96.53&68.47&25.02\\
    $\alpha=1$  &97.44&97.83&97.21&96.69&68.61&25.11\\
    $\alpha=2$  & 98.10 & 98.23 & 97.46& 96.84 &68.83&25.24\\
    $\alpha=3$  &97.94&98.12&97.37&96.79&68.77&25.20\\
    \bottomrule
\end{tabular}}
\caption{Ablation study for the weighting factor $\alpha$.}
\label{tab:a}
\end{table}

\subsection{Impact of the Key-Frame Selection Measures}
In LongDiff, we use the combination of two measures, image entropy and frame differencing, to select informative key frames by comparing the following variants: 1) \textbf{w/o Entropy}, where only image entropy is used as the sole measure to select key frames. 2) \textbf{w/o Differencing}, where frames are selected solely based on the frame differencing measure. As shown in \cref{tab:mesures}, the combination of these two measures yields the best results. In addtion, both variants outperform previous training-free methods\cite{lu2024freelong, qiu2023freenoise}.  
\begin{table}[h]
\centering
\resizebox{\linewidth}{!}{
\begin{tabular}{lcccccc}
    \toprule
    Method & SC $\uparrow$ & BC $\uparrow$ & MS $\uparrow$  & TF $\uparrow$ & IQ $\uparrow$&OC $\uparrow$  \\
    \midrule
    w/o Entropy & 97.10&97.58&96.92&96.53&68.47&25.02\\
    w/o Differencing  &97.35&97.75&97.06&96.61&68.50&25.02\\
    \midrule
    Ours & 98.10 & 98.23 & 97.46& 96.84 &68.83&25.24\\
    \bottomrule
\end{tabular}}
\caption{Ablation study for the key-frame selection measures.}
\label{tab:mesures}
\end{table}

\subsection{Impact of the Proportion of LongDiff Modules}
In our main experiments, we uniformly replace 50\% of the temporal attention layers in the short video model with our LongDiff modules. Here, we explore the impact of varying the proportion of the number of replaced temporal attention layers with LongDiff modules. As shown in \cref{tab:weight}, the performance improves noticeably when the proportion of LongDiff is below 50\%, and the improvement trend plateaus beyond this point.  Based on this observation, we choose to uniformly replace 50\% of the temporal attention layers with our LongDiff modules to achieve good results while maintaining efficiency.

\begin{table}[h]
\centering
\resizebox{\linewidth}{!}{
\begin{tabular}{lcccccc}
    \toprule
    Method & SC $\uparrow$ & BC $\uparrow$ & MS $\uparrow$  & TF $\uparrow$ & IQ $\uparrow$&OC $\uparrow$  \\
    \midrule
    Proportion 25.0\%  &  94.86&96.14&96.07&95.30&66.53&23.94\\
    Proportion 50.0\%  & 98.10 & 98.23 & 97.46& 96.84 &68.83&25.24\\
    Proportion 75.0\%  & 98.40 & 98.51 & 97.63& 96.98 &68.92&25.25\\
    \bottomrule
\end{tabular}}
\caption{Ablation study for the proportion of LongDiff modules.}
\label{tab:weight}
\end{table}

\begin{figure*}[t]
    \centering\textbf{}
    \includegraphics[width=0.85\linewidth]{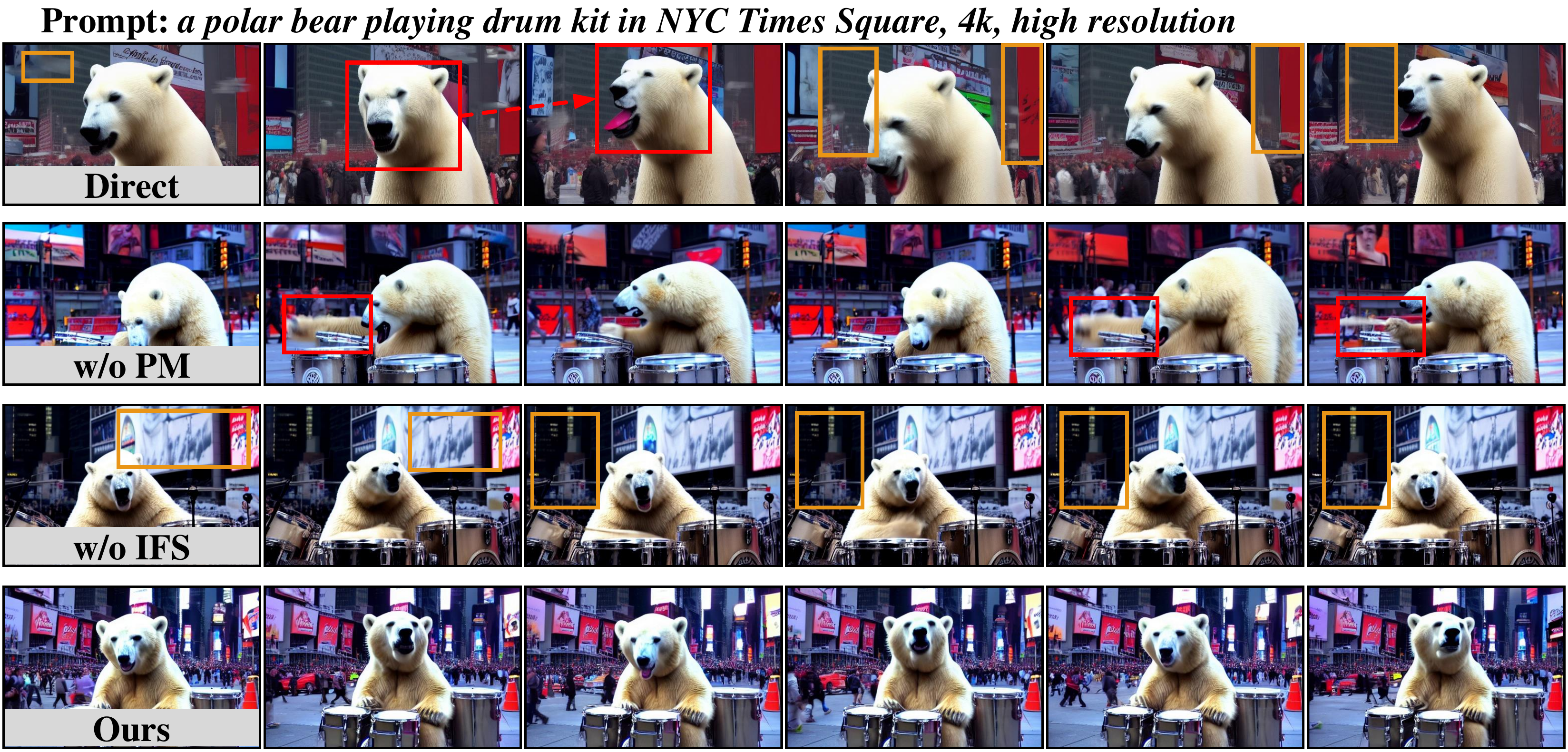}
    \vspace{-0.2cm}
     \caption{\textbf{Ablation Study of the Main Components of LongDiff.} Here, we show the qualitative comparison of LongDiff with two variants: \textbf{Ours (w/o PM)}, where we remove the \texttt{GROUP} and \texttt{SHIFT} operations and thus use the original relative positions to compute temporal attention. 2) \textbf{Ours (w/o IFS)}, where we remove the IFS mask in temporal attention computation, requiring each query frame to correlate with all frames for information passing during video generation.
     As shown, videos generated from the (w/o PM) variant  exhibit abrupt temporal transition between frames, particularly noticeable in the bear's hand. On the other hand, videos generated from the (w/o IFS) variant lack some visual details, manifesting as as a blurry ``NYC Times Square". 
     We illustrate inferior temporal consistency and the the visual detail issues using \textcolor{red}{red} and \textcolor{orange}{orange} boxes, respectively. 
     }
 \label{fig:ablation}
 \vspace{-0.2cm}
\end{figure*}

\section{More Qualitative Results}
\begin{figure*}[]
    \centering
    \includegraphics[width=0.9\linewidth]{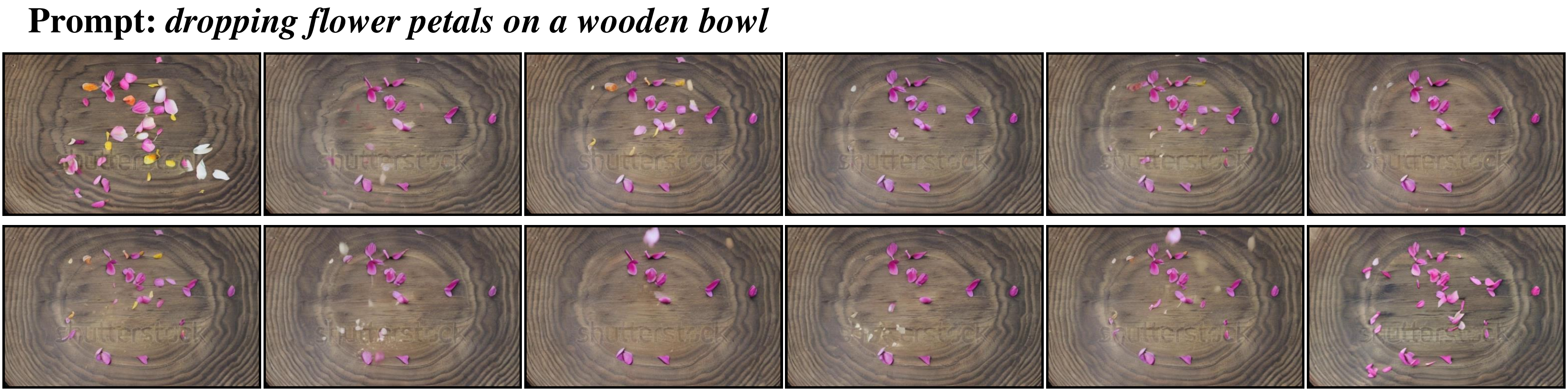}
    \vspace{-0.3cm}
     \caption{\textbf{Longer Video Generation.} Here, we equip VideoCrafter\cite{chen2023videocrafter1} with our LongDiff to generate 256-frame videos. As shown, these generated videos maintain  temporal consistency and visual details. This further demonstrates the efficacy of out method.}
 \label{fig:longer}
 \vspace{-0.2cm}
\end{figure*}

\begin{figure*}[]
    \centering
    \includegraphics[width=1\linewidth]{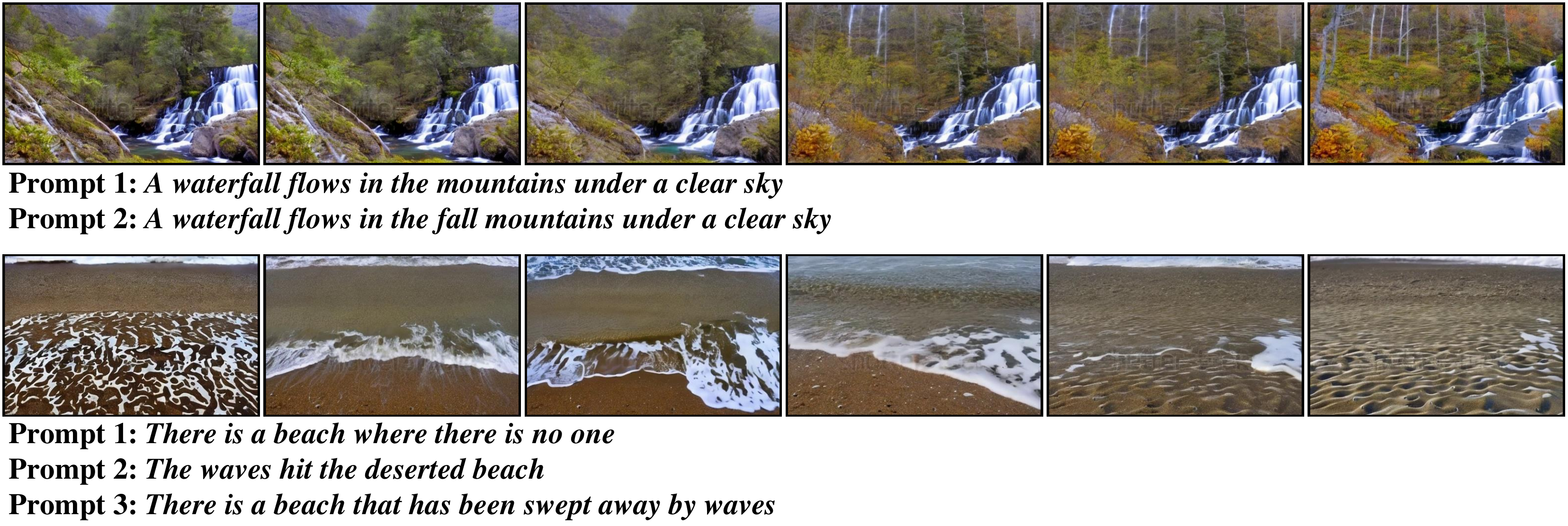}
    \vspace{-0.3cm}
     \caption{\textbf{Multi-Prompt Video Generation.} Our LongDiff can be easily adapted for multi-prompt video generation by assigning distinct prompts to each video segment following \cite{lu2024freelong, qiu2023freenoise}. As shown,  the output of our LongDiff maintains temporal consistency and visual details across different segments.}
 \label{fig:multi}
\end{figure*}

\begin{figure*}[h]
    \centering
     \vspace{-0.5cm}
    \includegraphics[width=\linewidth]{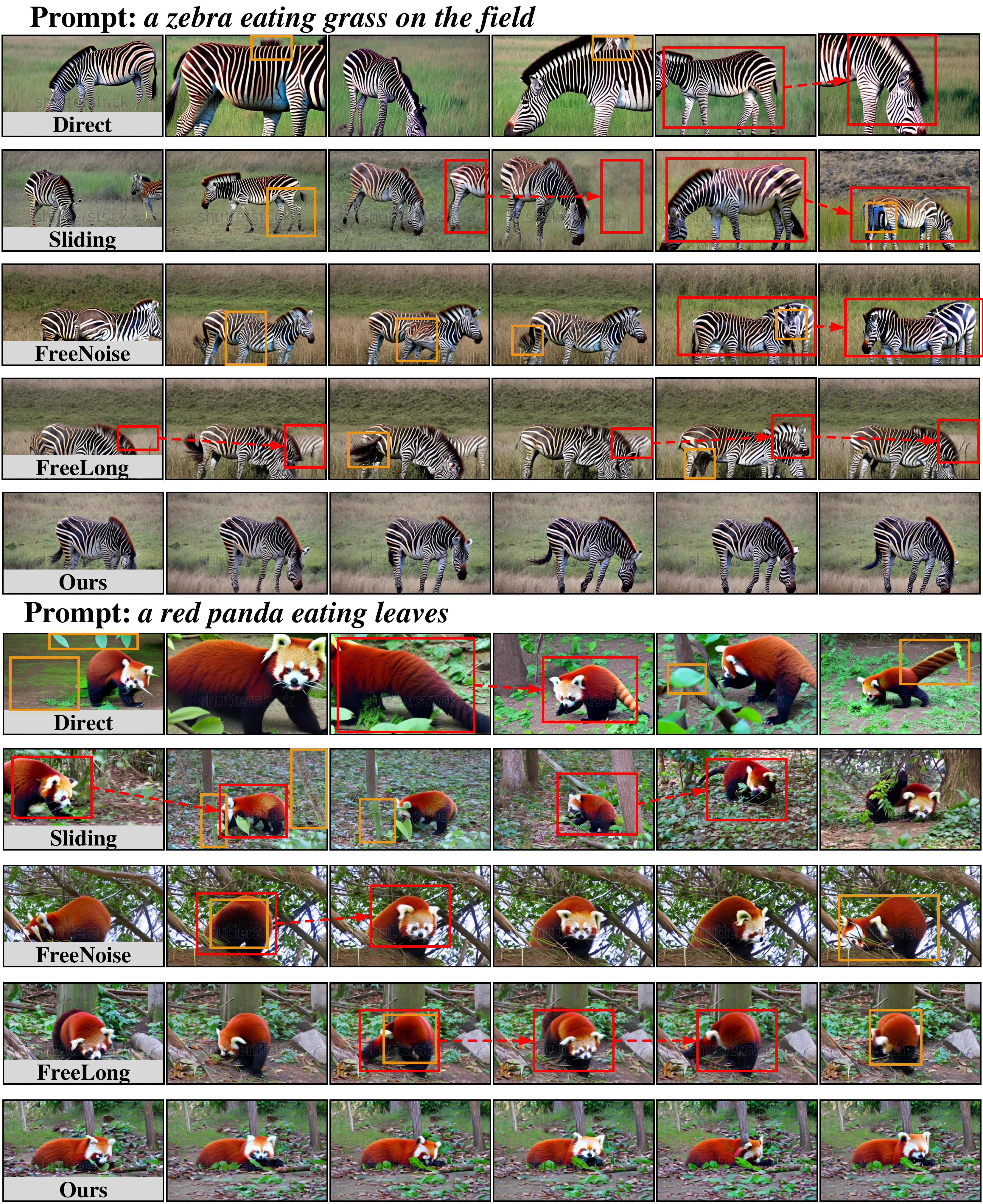}
     \caption{Qualitative comparisons of long video generation (128 frames) based on VideoCrafter\cite{chen2023videocrafter1}. 
    Compared to our LongDiff, videos generated by other methods lack temporal consistency to some extent (e.g., zebras that suddenly appear and disappear in the videos generated from the first prompt; drastic motion changes of the red panda in the videos generated from the second prompt), and suffer from visual detail issues (e.g., blurred zebra bodies in the videos generated from the first prompt; fuzzy leaves and red pandas in the videos generated from the second prompt). We illustrate inferior temporal consistency and visual detail issues using \textcolor{red}{red} and \textcolor{orange}{orange} boxes, respectively.
     }
 \label{fig:VC}
\end{figure*}

\begin{figure*}[h]
    \centering
    \vspace{-0.5cm}
    \includegraphics[width=1\linewidth]{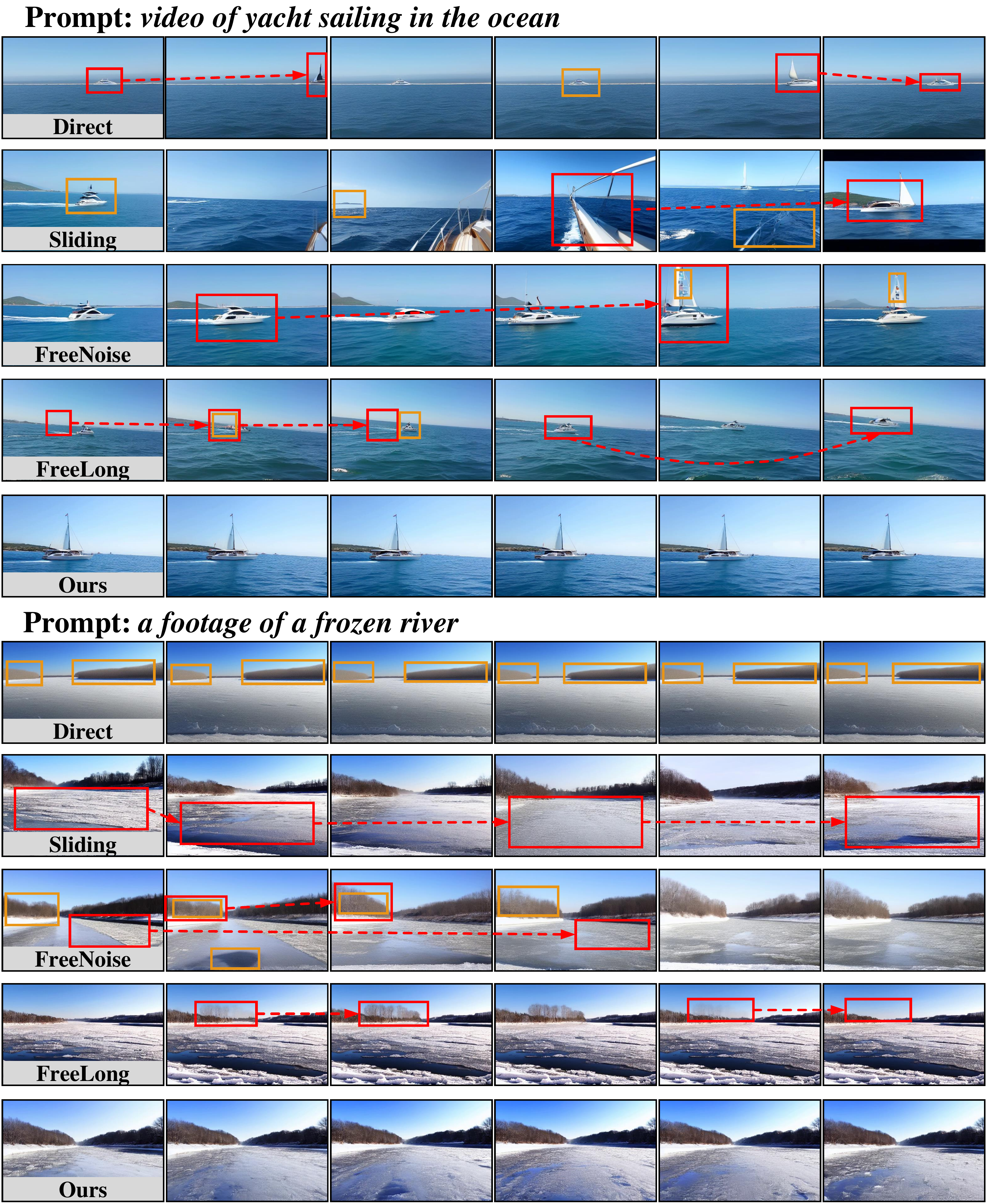}
     \caption{Qualitative comparisons of long video generation (128 frames) based on LaVie\cite{wang2023lavie}. 
    Compared to our LongDiff, videos generated by other methods lack temporal consistency to some extent (e.g., altered yacht structures in the videos generated from the first prompt; changing river surfaces and trees that suddenly appear and disappear in the videos generated from the second prompt), and suffer from visual detail issues (e.g., the fuzzy yacht in the videos generated from the first prompt; the blurred forest in the videos generated from the second prompt). We illustrate inferior temporal consistency and the visual detail issues using \textcolor{red}{red} and \textcolor{orange}{orange} boxes, respectively. 
     }
 \label{fig:lavie}
\end{figure*}

In this section, we provide more qualitative results regarding the ablation study of the main components of LongDiff (see \cref{fig:ablation}), longer video generation (see \cref{fig:longer}), multi-prompt video generation (see \cref{fig:multi}), and more generated videos (see \cref{fig:VC} and \cref{fig:lavie}).

\section{Proofs}
\subsection{Detailed Proof of Theorem 1}
Here, we provide a detailed proof of Theorem 1, which is mainly based on \cite{han2024lm}. 
For ease of reading, we restate Theorem 1 from the main paper below.

\begin{theorem}
    \label{the1}
Define the attention logit function in temporal attention as \(f(\mathbf{q}, \mathbf{k}, p)\),
which maps the query frame \(\mathbf{q}\), key frame \(\mathbf{k}\), and their relative position $p$ to a scalar value. 
Consider a video generation task with $N$ frames, where the model categorizes the $2N$$-$$1$ relative positions into \(g(N)\) groups.
Here, \(g(N) \in \mathbb{N}\) is a non-decreasing and unbounded function representing the model's capability to differentiate relative positions. Additionally, assume that any two relative positions \(p\) and \(p'\) within the same group satisfy \(d_f(p, p') \leq \epsilon\), where \(d_f\) is the distance function associated with the attention logit function \(f\). Then the following  holds:
\begin{equation}
\label{eq:sup}
\sup_{-(N-1)\leq p \leq N-1} |f(\mathbf{q}, \mathbf{k}, p)| \geq \left(\frac{g(N)}{2}\right)^{\frac{1}{2r}}\frac{ \epsilon}{4e}
\end{equation}
where $r$ is the pseudo-dimension of the function class $\mathcal{H} = \{f(\cdot, \cdot, p) \mid p \in \mathbb{Z}\}$, and $e$ is the Euler's number. 
\end{theorem}
The distance function $d_f$ can be rewritten in a more detailed form as:
\begin{equation}
\label{eq:distance}
d_f(p, p') = \mathbb{E}_{\mathbf{q\sim Q, k\sim K}}(f(\mathbf{q},\mathbf{k},p)-f(\mathbf{q},\mathbf{k},p'))^2
\end{equation}
where $\mathbf{Q}$ and $\mathbf{K}$ are the trained distributions for $\mathbf{q}$ and $\mathbf{k}$. 
To assist in proving the inequality in Theorem 1, the following lemma is introduced from \cite{haussler1992decision}.

\begin{lemma}
\label{lemma}
    Let $\mathcal{H}=\{h(z)\}$ be a family of functions that map a set $\mathbf{Z}$ into $[0, M]$ with pseudo-dimension $\dim_{P}(\mathcal{H}) = r$, where $1 \leq r < \infty$. Let $P$ be a probability measure on $\mathbf{Z}$. Then, for all $0 < \epsilon \leq M$, the $\epsilon$-cover of $\mathcal{H}$ under the metric $d(h_1, h_2) =\mathbb{E}_{z\sim P} (h_1(z) - h_2(z))^2$ is bounded by:
\begin{equation}
\mathcal{N}_P(\epsilon, \mathcal{H}, d) \leq 2 \left( \frac{2eM}{\epsilon} \ln \frac{2eM}{\epsilon} \right)^r
\end{equation}
where $N_P(\epsilon, \mathcal{H}, d)$ is the cover size, defined as the smallest cardinal of a cover-set \( \mathcal{H}' \) such that for every entry \( h \in \mathcal{H} \), there exists at least one entry \( h' \in \mathcal{H}' \) within \( \epsilon \) distance from \( h \).
\end{lemma}    
Based on \cref{lemma}, Theorem 1 can be proven by contradiction as follows.
\begin{proof}
First, let the negation of \cref{eq:sup} in Theorem 1 be assumed to hold:
\begin{equation}
\label{eq:assumation}
\sup_{-(N-1)\leq p \leq N-1} |f(\mathbf{q}, \mathbf{k}, p)| < \left( \frac{g(N)}{2} \right)^{\frac{1}{2r}} \frac{\epsilon}{4e}  =  a
\end{equation}
This indicates that the function family $\mathcal{H} = \{f(\cdot, \cdot, p) | p \in \mathbb{Z} \}$ maps the input to the range $[-a, a]$. Without loss of generality, all values from the range $[-a,a]$  can be shifted to the range $[0, 2a]$  to apply \cref{lemma}. Then, according to \cref{lemma}\footnote{A prerequisite for applying \cref{lemma} is that $\mathcal{H}$ has a bounded pseudo-dimension $\dim_{P}(\mathcal{H}) = r$ (i.e., $1\leq r <+\infty$). 
It will be shown that $\mathcal{H}$ satisfies this prerequisite later.}, the $\epsilon$-cover size $\mathcal{N}_P(\epsilon, \mathcal{H}, d_f)$ of $\mathcal{H}$ satisfies that:
\begin{equation}
\label{proof1}
    \mathcal{N}_P(\epsilon, \mathcal{H}, d_f)  \leq 2 \left( \frac{4ea}{\epsilon} \ln \frac{4ea}{\epsilon} \right)^r
\end{equation}
By substituting $a = \left( \frac{g(N)}{2}\right)^{\frac{1}{2r}} \frac{\epsilon}{4e}$ (defined in \cref{eq:assumation}) into \cref{proof1}, the following expression is obtained:
\begin{equation}
\begin{aligned}
\mathcal{N}_P(\epsilon, \mathcal{H}, d_f)  &\leq 2 \Bigg(\left( \frac{g(N)}{2}\right)^{\frac{1}{2r}}  \ln\left( \frac{g(N)}{2}\right)^{\frac{1}{2r}} \Bigg)^{r}  \\
& < 2 \Bigg(\left( \frac{g(N)}{2}\right)^{\frac{1}{2r}} \left( \frac{g(N)}{2}\right)^{\frac{1}{2r}} \Bigg)^{r} = g(N)
\end{aligned}
\end{equation}
This indicates that, if the assumption in \cref{eq:assumation} holds, the $\epsilon$-cover size $\mathcal{N}_P(\epsilon, \mathcal{H}, d_f)$ is smaller than $g(N)$.
In other words, we cannot find $g(N)$ distinct functions in the function family $\mathcal{H} = \{f(\cdot, \cdot, p) \mid p \in \mathbb{Z} \}$ such that the pairwise distances (measured by $d_f$) between them are greater than $\epsilon$. This implies that the number of distinct relative positions differentiated by the model is less than $g(N)$, which contradicts the definition of $g(N)$. Therefore, \cref{eq:assumation} does not hold, and thus \cref{eq:sup} in Theorem 1 is proven.
\end{proof}

\noindent \textbf{Pseudo-Dimension of $\mathcal{H}$}. 
As discussed above, \cref{lemma} is introduced to assist in proving \cref{the1}, which requires that $\mathcal{H}$ has a bounded pseudo-dimension $\dim_{P}(\mathcal{H}) = r$. 
Notably, $\mathcal{H} = \{f(\cdot, \cdot, p) \mid p \in \mathbb{Z} \}$ represents the family of attention logit functions, whose form varies depending on the RPE mechanisms. For RoPE, the logit function \( f(\cdot, \cdot, p) \) can be expressed as a weighted sum of a finite set of sinusoidal functions \(\{\sin(\omega_i p), \cos(\omega_i p)\}\), where the size of this set equals the feature dimension \(k\). Based on the properties of pseudo-dimensions, it follows that \(\dim_P(\mathcal{H}_1 + H_2) \leq \dim_P(\mathcal{H}_1) + \dim_P(\mathcal{H}_2)\), and the pseudo-dimension of scaling a single function is at most 2. Therefore, the pseudo-dimension of the whole family is bounded by \(\dim_P(\mathcal{H}) \leq 2k\), which satisfies the requirement in \cref{lemma}.

\noindent\textbf{Analysis of Theorem 1.} \cref{the1} implies that, the ability of a video model to distinguish between different relative positions is constrained by the supremum of the model's temporal attention logits. Building on \cref{the1}, here, we further analyze whether existing video models can accurately identify frame order during long video generation. Recall that for a video model to correctly identify frame order in a video of length $N$, it must be capable of distinguishing between $2N - 1$ distinct relative positions using its temporal attention logits. According to \cref{the1}, this requirement means that a video model capable of correctly identifying frame order must satisfy \cref{eq:sup} when $g(N)$ is set to $2N - 1$. Conversely, if the inequality in \cref{eq:sup} fails to hold for $g(N) = 2N - 1$, it suggests that the supremum of the temporal attention logits is inadequate for the model to handle $2N - 1$ distinct positions. Consequently, the model is unable to correctly identify frame order. Based on the above arguments, here, we perform our analysis taking the LaVie \cite{wang2023lavie} video model as a case study, and use it to directly generate 128-frame (i.e., $N=128$) videos. 
Notably, as shown in \cref{eq:sup}, to compute it, we need to determine the values of $r$ and $\epsilon$. Below, we then first discuss how we determine the values of $r$ and $\epsilon$ for LaVie in our analysis.

Specifically, w.r.t. $r$, in LaVie, RoPE \cite{su2024roformer} is employed as the RPE mechanism, and only 32 dimensions (i.e., $k=32$) of the query and key features are processed by RoPE in each attention head. Additionally, as discussed earlier, for models using RoPE as the RPE mechanism, $r \leq 2k$ (i.e., $r\leq 64$). Notably, the right-hand side of \cref{eq:sup} is negatively correlated with $r$. Hence, if $r=64$ causes the inequality to fail, then the inequality does not hold for any $r< 64$. 
We then set $r=64$ for the subsequent analysis here.

Meanwhile, to determine the value of $\epsilon$, we first examine a scenario where these \(2N - 1\) positions are uniformly distributed to $2N-1$ groups (given $g(N)=2N-1$). And the boundaries of these clusters (groups)  are precisely situated in the middle of two adjacent positions. Consequently, the maximum intra-cluster (group) distance for the cluster that includes position \(p\)  can be determined by calculating \(d_f(p - 0.5, p + 0.5)\). 
According to the definition in Theorem \ref{the1}, \(\epsilon\) is greater than the maximum intra-cluster  distance (measured by  \(d_f\)) across all position clusters. 
With the maximum intra-distance of each group, we can determine the lower bound of \(\epsilon\), denoted as \(\Omega(\epsilon_{\text{uni}})\). Notably, though \(\Omega(\epsilon_{\text{uni}})\) is obtained based on the assumption that these \(2N-1\) positions are uniformly clustered, for any non-uniformly distributed scenarios, there must exist at least one position cluster of larger size with a greater maximum intra-cluster distance.
This means the true lower bound of \(\epsilon\) is greater than \(\Omega(\epsilon_{\text{uni}})\). 
Additionally, the right-hand side of \cref{eq:sup} is positively correlated with $\epsilon$. Hence, if setting $\epsilon$ to $\Omega(\epsilon_{\text{uni}})$ causes the \cref{eq:sup} to fail, then the inequality does not hold for any $\epsilon$. Thus, we here set \(\epsilon = \Omega(\epsilon_{\text{uni}})\) for subsequent analysis.

After determining the values of $r$ and $\epsilon$, we extract query and key features from all the temporal attention heads to compute both the left and right sides of \cref{eq:sup}. We find that when generating 128-frame videos, only query and key features in 40\% of attention heads satisfy the inequality in \cref{eq:sup}, and this percentage decreases to 34\% when setting $N=256$. 
This suggests that the supremum of the temporal attention logits is insufficient for the model to achieve $g(N) = 2N - 1$. In other words, the existing video model can struggle in identifying correct frame order.

\subsection{Detailed Proof of Theorem 2}

Here, we provide a detailed proof of Theorem 2. For ease of reading, we restate Theorem 2 from the main paper below.
\begin{theorem}
\label{t2}
    When generating a video with $N$ frames, the information entropy $H$ of temporal correlations over frames of the video sequence, is lower bounded by~\cite{han2024lm}:
\begin{equation}
\setlength{\abovedisplayskip}{3pt}
\setlength{\belowdisplayskip}{3pt}
    \label{eq:entropy}
    H\left(\frac{e^{a_i}}{\Sigma^{N}_{j=1}e^{a_j}}|1\leq i\leq N\right)  \geq \ln N -2B,
\end{equation}
where $\{a_i\}_{i=1}^N$ are the attention logits with boundary $[-B, B]$.
\end{theorem}
\begin{proof}
The information entropy $H$ of a discrete distribution $P$ is given as $H(P) = - \Sigma_i p_i \ln p_i $. Hence, the information entropy of temporal correlation is computed as follows~\cite{han2024lm}:

\begin{equation}
\begin{aligned}
& H\left(\frac{e^{a_i}}{\Sigma^{N}_{j=1}e^{a_j}}|1\leq i\leq N\right) \\
&\qquad  \qquad= -\sum_i \frac{e^{a_i}}{\sum_j e^{a_j}} \ln \frac{e^{a_i}}{\sum_j e^{a_j}} \\
&\qquad  \qquad= -\sum_i \frac{e^{a_i}}{\sum_j e^{a_j}} \left(a_i - \ln \sum_j e^{a_j} \right) \\
&\qquad  \qquad= -\sum_i \frac{e^{a_i}}{\sum_j e^{a_j}} a_i + \ln \sum_j e^{a_j} \\
&\qquad  \qquad\geq -\max_i a_i + \ln (N e^{-B}) \\
&\qquad  \qquad\geq \ln N - 2B \\
\end{aligned}
\end{equation}
\end{proof}

\section{More Experiment Results}

\subsection{User Study}
Following \cite{qiu2023freenoise}, we carried out a user study to assess our results based on human subjective judgment. In this study, participants were shown generated long videos using LaVie as the short video model from all methods (a total of 250 videos), with the examples presented in a random order to eliminate potential bias. Participants were then asked to score the generated videos on a scale of 1 to 5 according to three evaluation criteria: content consistency, video quality, and video-text alignment. The average scores for each method are reported in \cref{tab:user}.
As shown, our method received the highest ratings across all metrics.

\begin{table}[h]
\centering
\resizebox{\linewidth}{!}{
\begin{tabular}{lccc}
    \toprule
    Method & Content Consistency $\uparrow$& Video Quality & Video-Text Alignment  \\
    \midrule
    Direct  &   2.8  &1.9 & 2.3 \\
    Sliding  & 1.8  &   3.1& 2.5\\ 
    FreeNoise~\cite{qiu2023freenoise}  &  3.3  &3.6&  3.5 \\  
    FreeLong~\cite{lu2024freelong}  & 3.7   & 3.8&3.9  \\
    \midrule 
    Ours &  4.7  &   4.6    & 4.7\\
    \bottomrule
\end{tabular}}
\caption{Comparison based on user study.}
\label{tab:user}
\end{table}

\subsection{Inference Time}

\setlength{\columnsep}{0.05in}
\begin{wraptable}[8]{r}{0.5\columnwidth}
\vspace{-0.4cm}
\centering
\resizebox{\linewidth}{!}{
\begin{tabular}{lc}
    \toprule
    Method & Inference Time $\downarrow$ \\
    \midrule
    Direct  &  4.0s\\
    Sliding  &  5.4s\\ 
    FreeNoise~\cite{qiu2023freenoise}  &5.4s\\  
    FreeLong~\cite{lu2024freelong}  &  4.7s\\
    \midrule
    Ours & 5.5s\\
    \bottomrule
\end{tabular}}
\vspace{-0.3cm}
\caption{Comparison of inference time.}
\label{tab:time}
\end{wraptable}

In this section, we compare the inference times (time required for each denoising step) of our LongDiff with other training-free methods \cite{qiu2023freenoise, lu2024freelong} and two basic methods, \textbf{Direct} and \textbf{Sliding}, on the NVIDIA A6000 Ada GPU. We apply all methods to LaVie and generate 128-frame videos for comparison. As shown in \cref{tab:time}, Our LongDiff significantly improves the quality of long videos generated by the short video model and achieves state-of-the-art results with only a modest increase in inference time compared to the Direct method.

\subsection{Evaluation on Video Models with Absolute Positional Encoding}

Our LongDiff can also be adapted to video models utilizing absolute positional encoding mechanisms, such as sinusoidal position encoding \cite{vaswani2017attention}. This is achieved by performing the \texttt{GROUP} and \texttt{SHIFT} operations directly on the frame position rather than the relative positions among frames. Here, we take Animatediff \cite{guo2023animatediff}, which uses sinusoidal position encoding for temporal attention computation, as a case study to evaluate the efficacy of our LongDiff. 
Specifically, we adapt Animatediff to generate 128-frame long videos with a resolution of 255$\times$255.
As shown in \cref{tab:animate}, compared to other training-free methods, LongDiff achieves the best performance across all the metrics.

\begin{table}[h!]
\centering
\resizebox{\linewidth}{!}{
\begin{tabular}{lcccccc}
    \toprule
    Method & SC $\uparrow$ & BC $\uparrow$ & MS $\uparrow$  & TF $\uparrow$ & IQ $\uparrow$&OC $\uparrow$  \\
    \midrule
    Direct  &  92.25& 94.35& 97.42& 96.75& 49.27& 20.01\\
    Sliding  &  86.62& 92.68& 97.86& 96.95& 60.51& 23.42\\
    FreeNoise~\cite{qiu2023freenoise}  &  95.84& 96.75& 98.92& 98.61& 64.69& 24.78\\
    FreeLong~\cite{lu2024freelong}  &  95.11& 95.86& 97.72& 98.10& 60.23& 23.51\\
    \midrule
    Ours  &  97.54& 97.39& 98.98& 98.70& 65.14& 25.11\\
    \bottomrule
\end{tabular}}
\caption{Quantitative comparisons of longer video generation (128
frames) on the Animatediff.} 
\label{tab:animate}
\end{table}

\end{document}